\documentclass{egpubl}
\usepackage{float}
\usepackage{array}
\newcolumntype{H}{>{\setbox0=\hbox\bgroup}c<{\egroup}@{}}
\usepackage{amsmath}
\usepackage{multirow}
\usepackage{subcaption}

\usepackage[dvipsnames]{xcolor}

\newcommand{\rev}[1]{{\color{red}#1}}

\definecolor{turquoise}{cmyk}{0.65,0,0.1,0.3}
\definecolor{purple}{rgb}{0.65,0,0.65}
\definecolor{dark_green}{rgb}{0, 0.5, 0}
\definecolor{orange}{rgb}{0.8, 0.6, 0.2}
\definecolor{red}{rgb}{0.8, 0.2, 0.2}
\definecolor{darkred}{rgb}{0.6, 0.1, 0.05}
\definecolor{blueish}{rgb}{0.0, 0.3, .6}
\definecolor{light_gray}{rgb}{0.7, 0.7, .7}
\definecolor{pink}{rgb}{1, 0, 1}
\definecolor{greyblue}{rgb}{0.25, 0.25, 1}

\definecolor{teal}{rgb}{0, 0.39, 0.39 }
\definecolor{brown}{rgb}{0.6, 0.3, 0.39 }
\newcommand{\secondbest}[1]{{\underline{#1}}}
\newcommand{\best}[1]{\mathbf{{#1}}}

\definecolor{red}{rgb}{0, 0, 0}

\newcommand{\modelname}{LEAD\xspace}
\newcommand{\titlename}{LEAD: Latent Realignment for Human Motion Diffusion\xspace}

\newcommand{\diffnet}{\epsilon_{\theta}}

\newcommand{\vaeE}{E_{vae}}
\newcommand{\vaeD}{D_{vae}}
\newcommand{\projE}{E_{proj}}
\newcommand{\projD}{D_{proj}}
\newcommand{\vaeZ}{z^{vae}}
\newcommand{\projZ}{z^{proj}}

\newcommand{\st}{}
\renewcommand\st[1]{}
\newcommand{\rmstd}{}
\renewcommand\rmstd[1]{}

\makeatletter
\DeclareRobustCommand\onedot{\futurelet\@let@token\@onedot}
\def\@onedot{\ifx\@let@token.\else.\null\fi\xspace}

\def\eg{\emph{e.g}\onedot} 
\def\ie{\emph{i.e}\onedot}

\def\etal{\emph{et al}\onedot}
\makeatother

\usepackage{xspace}
\usepackage{amssymb}
\usepackage{multirow}
\usepackage{booktabs}

 \JournalSubmission    % uncomment for submission to Computer Graphics Forum
 \electronicVersion % can be used both for the printed and electronic version

\ifpdf \usepackage[pdftex]{graphicx} \pdfcompresslevel=9
\else \usepackage[dvips]{graphicx} \fi

\PrintedOrElectronic

\usepackage{t1enc,dfadobe}

\usepackage{egweblnk}
\usepackage{cite}

\title[\titlename]{\titlename}

\author[N. Andreou]
{\parbox{\textwidth}{\centering N. Andreou$^{1,2,3}$\thanks{Work done prior to joining Amazon. \\ Corresponding author: nefeliandreou@outlook.com},
         X. Wang$^{2}$,  
         V. Fernández Abrevaya$^{3}$, 
         M.P. Cani$^{2}$,
         Y. Chrysanthou$^{4}$ and
         V. Kalogeiton$^{2}$
        }
        \\
{\parbox{\textwidth}{\centering $^1$University of Cyprus, Cyprus\\
         $^2$LIX, Ecole Polytechnique, CNRS, Institut Polytechnique de Paris, France\\
         $^3$Max Planck Institute for Intelligent Systems, Tübingen, Germany\\
         $^4$CYENS - Centre of Excellence, Cyprus
       }
}
}

\begin{document}

% uncomment for using teaser
\teaser{
 \includegraphics[width=\linewidth]{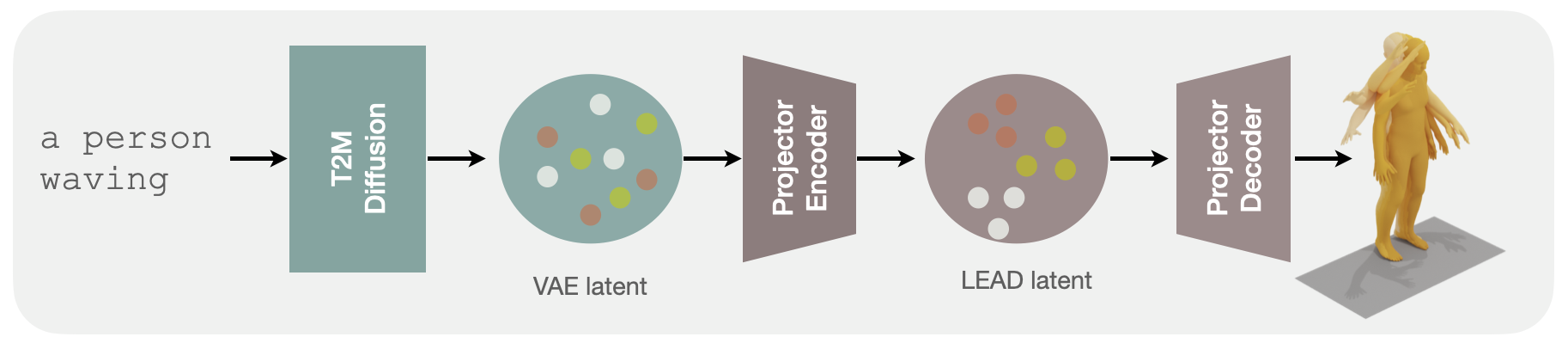}
 \centering
  \caption{\modelname enhances human motion diffusion models by introducing a latent realignment scheme that enforces semantic consistency in the latent space. We show that this leads to improved realism and expressiveness in the tasks of motion generation and motion inversion.}
\label{fig:teaser}
}

\maketitle
%-------------------------------------------------------------------------
\begin{abstract}
Our goal is to generate realistic human motion from natural language. 
Modern methods often face a trade-off between model expressiveness and text-to-motion alignment. Some align text and motion latent spaces but sacrifice expressiveness; others rely on diffusion models producing impressive motions, but lacking semantic meaning in their latent space. This may compromise realism, diversity, and applicability. 
Here, we address this by combining latent diffusion with a realignment mechanism, producing a novel, semantically structured space that encodes the semantics of language. Leveraging this capability, we introduce the task of textual motion inversion to capture novel motion concepts from a few examples.
For motion synthesis, we evaluate \modelname on HumanML3D and KIT-ML and show comparable performance to the state-of-the-art in terms of realism, diversity, and text-motion consistency. 
Our qualitative analysis and user study reveal that our synthesized motions are sharper, more human-like and comply better with the text compared to modern methods. For motion textual inversion, our method demonstrates improved capacity in capturing out-of-distribution characteristics in comparison to traditional VAEs.


\end{abstract}

\section{Introduction}
\label{sec:intro}

%% --- Goal
Text-to-motion (T2M) generation is the process of creating human-like motion that reflects a given language instruction. Generating motions that comply with textual descriptions is a task that received significant attention~\cite{Chen:2023, Tevet:2022, Petrovich:2022, Athanasiou:2022} due to its potential to democratize 3D content creation and its numerous applications in fields such as robotics~\cite{Plappert:2018}, entertainment~\cite{Holden:2016,Holden:2017} and virtual reality~\cite{Bhattacharya:2021}.

%% --- Problem
A text-to-motion model should be able to accurately reproduce arbitrary descriptions in natural language while accounting for the many-to-many nature of the problem. 
This is a challenging task, since there is a large discrepancy between the space of natural language and the space of human motions (\ie skeletal poses)~\cite{Tevet:2022:MotionCLIP, Ahuja:2019, Chen:2023}.
Solutions can be classified into two main categories. 
Initial works such as TEMOS~\cite{Petrovich:2022}, L2JP~\cite{Ahuja:2019} and MotionCLIP~\cite{Tevet:2022:MotionCLIP} build a common latent space that simultaneously encodes natural language and motion. These approaches typically rely on (variational) autoencoders~\cite{Kingma:2013:VAE} and are thus restricted in the distribution they can model, where the mapping from text to motion is either assumed to be one-to-one~\cite{Ghosh:2017, Ahuja:2019} or to follow a normal distribution in the latent space \cite{Petrovich:2022}. 

A second category leverages diffusion models to learn the distribution of human motions conditioned on text, offering enhanced diversity and realism surpassing the previous constraints~\cite{Chen:2023, Kim:2022,Zhang:2022,Tevet:2022}. Despite their impressive generations, we observe that the synthesized animations might not consistently follow the input text, and may still generate unrealistic movement. We hypothesize that a semantically structured motion latent space, \ie one that inherits some of the rich properties of the language space, can facilitate and improve the task of text-to-motion generation. 

In the image domain, diffusion models have gone beyond general text-to-image generation towards \emph{personalized} generation, 
where the goal is to synthesize concepts that are hard to describe (\eg. a specific instance of an object) based on a few examples. The work of Gal \etal~\cite{Gal:2022} was among the first to address this through an optimization approach, where the unknown embedding in the language space is learnt given a set of examples. This \textit{inversion} approach that recovers the embedding of a concept could certainly be beneficial for the motion domain, too. 
If one could invert a motion diffusion model, the generated motions could then be customized to reflect specific attributes of the exemplar motions such that the generated sequence is not only contextually relevant to the given text, but also aligned with the unique characteristics of the references.

Nevertheless, to the best of our knowledge, no motion diffusion work demonstrates the ability to perform inversion or personalization. We argue (and our experiments suggest) that stems from the misalignment between the latent space in which they operate and the latent space of a language model.

%% --- Our approach
In this work, we propose \modelname, a new text-to-motion model based on latent diffusion~\cite{Roombach:2022, Chen:2023} that addresses the lack of semantic structure in the latent space. Key to our approach is the incorporation of a \emph{projector} module that realigns the original latent space of a motion VAE towards one that is in more congruence with a language model, namely CLIP~\cite{Radford:2021}. The proposed module is realized via an autoencoder that efficiently projects the diffused latents into the new space, and re-projects them back into the VAE space at inference time. 
Furthermore, we introduce the paradigm of 
motion textual inversion (MTI), 
and observe that the proposed projector also improves this task. Similar to~\cite{Gal:2022}, we formulate motion inversion as the task of finding the embedding in the latent space of a pre-trained language model that best captures characteristic of the exemplar motion. 

We evaluate our method on the task of text-to-motion generation on the standard HumanML3D~\cite{Guo:2022} and KIT-ML~\cite{Mandery:2015:KIT} datasets. Our results show that \modelname achieves on-par performance to the state of the art in terms of motion quality while retaining good performance in terms of diversity and multimodality-- a trade-off that none of the competing methods can handle well. The effectiveness of \modelname is particularly evident for the task of personalized motion generation from text, with improved qualitative and quantitative results over a vanilla implementation of textual inversion on existing models.

Our contributions can be summarized as follows:
(a) We propose \modelname, a motion diffusion method relying on a text-motion realignment mechanism. 
(b) We show quantitatively and qualitatively that \modelname results in improved performance in two datasets for text-to-motion generation, without additional computational cost.
(c) We introduce the task of motion textual inversion and show that the simple yet elegant realignment paves the way for generating motions with specific characteristics from the text.
\section{Related Work}
\noindent
Human motion synthesis  can be split into unconditional and conditional. Unconditional synthesis~\cite{Yan:2019, Zhang:2020, Rui:2020} models the entire manifold of possible motions, while conditional synthesis introduces constraints such as audio or text that guide the generation. 
For a complete overview of motion synthesis, we refer the reader to~\cite{Mourot:2021, Kwiatkowski:2022y}. Here, we focus on conditional synthesis using multimodal constraints.

\noindent \textbf{Multimodal Motion Synthesis.}
To condition motion generation, research works have explored the use of
text~\cite{Petrovich:2021, Petrovich:2022, Guo:2020,Guo:2022}, images~\cite{Ghorbani:2020, Rempe:2021}, audio~\cite{Ginosar:2019}, music~\cite{Tseng:2023, Aristidou:2021, Lee:2019} and scenes~\cite{Hassan:2021, Starke:2021}. 
Generating motion from text is an intuitive way to produce 3D content and has received significant attention. Initial works targeted the action-to-motion task~\cite{Petrovich:2021, Guo:2020,Lucas:2022} that produces motions depicting a single action. 
% Guo~\etal~
\cite{Guo:2020} propose a temporal-VAE based on GRUs to produce diverse motions, with a disentangled representation to better capture the kinematic properties. \cite{Petrovich:2021} designs a transformer-based VAE with learnable tokens for each action, while \cite{Lucas:2022} compress motion into a discrete latent space and realize future states as next-index predictions. 

Instead of relying on a fixed set of action categories, several works incorporate a text-encoder that transforms natural language into a latent space that acts as the conditioning signal~\cite{Zhang:2023, Guo:2022:tm2t, Guo:2022, Bhattacharya:2021,jiang2024motiongpt}. 
Guo~\etal~\cite{Guo:2022} first learn motion codes using a motion autoencoder and then use a recurrent VAE to map the text condition to a motion snippet code sequence. Similarly, Zhang~\etal~\cite{Zhang:2023} learn a mapping from motion to discrete codes using a VQ-VAE~\cite{Oord:2021}, and use a transformer module to generate motion indices using text. MotionGPT~\cite{jiang2024motiongpt} extends this idea to a versatile framework designed to address a variety of motion-language related tasks such as motion<->text. 
Some text-to-motion works align the motion-text spaces directly~\cite{Ahuja:2019,Petrovich:2021,Petrovich:2022,Ghosh:2021,Athanasiou:2022}. JL2P~\cite{Ahuja:2019} learns a joint pose-language space with a cross-modal loss, while \cite{Ghosh:2021} follow a similar approach but construct two separate manifolds for the upper and lower body.
TEMOS~\cite{Petrovich:2022} bypasses the need for two manifolds and ensures diverse sampling by encoding distribution parameters using a VAE and a pre-trained language model. To enable sequential generation of motions, TEACH~\cite{Athanasiou:2022} augments the text-encoder branch of TEMOS to account for temporal compositions.

Recent works leverage the representational power of Large Language Models. SINC~\cite{Athanasiou:2023} enhances TEMOS~\cite{Petrovich:2021} by
exploiting GPT-3~\cite{GPT3} to distil knowledge on the correspondence between actions and body parts, allowing for spatial compositions from text. More related to our work, Tevet~\etal~\cite{Tevet:2022:MotionCLIP} propose MotionCLIP, a model that leverages the power of CLIP 

by aligning the latent motion representation to the image and text representations of CLIP~\cite{Radford:2021}.
Importantly, MotionCLIP~\cite{Tevet:2022:MotionCLIP} is designed to produce a one-to-one mapping between language and motion failing to properly reflect the inherent diversity and ambiguity of human action, where multiple distinct movements can be appropriate responses to the same verbal instruction.
Additionally, although MotionCLIP  
excels in action-to-motion and motion generation using high-level text, 

it does not account for global displacement 
and shows inconsistencies for out-of-distribution (OOD) generation. In a similar fashion, Ao~\etal~\cite{Ao:2023} demonstrate the power of LLMs and contrastive learning in creating a shared latent space between full-body gestures and speech semantics, which is used to improve the task of speech-to-gesture generation.

\noindent
\textbf{Motion Diffusion Models.}

Diffusion models have demonstrated unprecedented capabilities for text-to-image generation~\cite{Ho:2020, Roombach:2022}. Recently their potential has been explored for the T2M task~\cite{Tevet:2022, Zhang:2022, Kim:2022, Dabral:2022, Zhang:2023:ReMoDiffuse, Karunratanakul:2023,Wang:2023,xie2023omnicontrol}. 
Tevet~\etal~\cite{Tevet:2022} introduce a transformer-based motion diffusion model (MDM) and propose geometric losses specifically curated for the motion domain. 

MotionDiffuse~\cite{Zhang:2023} uses a cross-modality linear transformer as backbone to enable a soft control, thus increasing the diversity of generated motions. ReMoDiffuse~\cite{Zhang:2023:ReMoDiffuse} augments MotionDiffuse~\cite{Zhang:2023} with a retrieval mechanism that refines the denoising process. OmniControl~\cite{xie2023omnicontrol} introduces an additional spatial control signal for individual joints.
Dabral~\etal~\cite{Dabral:2022} design a system based on 1D U-Net with a cross-modal transformer and multi-head attention that can be conditioned on both textual and audio signals. More related to our work, inspired by image latent diffusion~\cite{Roombach:2022} Chen~\etal~\cite{Chen:2023} propose MLD, a method that performs diffusion over the VAE motion latent to handle noisy data and decrease computational complexity. MotionLCM~\cite{motionlcm} takes this a step further enabling faster inference through a motion latent consistency model. They first pre-train MLD and observe improved performance by replacing the CLIP~\cite{Radford:2021} text encoder with T5~\cite{t5}.
MotionLCM then distills information from the pre-trained MLD and uses a motion ControlNet in the latent motion space offering enhanced spatial control.

In our work, we explicitly align the motion latent space with language and demonstrate how a simple, yet elegant realignment step during inference can boost the realism in the generations, without sacrificing multimodality and diversity. In comparison to other controls which achieve similar degree of realism~\cite{xie2023omnicontrol}, our method generates realistic and expressive motions purely from language, without the need for explicit spatial control.

\noindent
\textbf{Textual Inversion in Diffusion Models.}
Textual inversion is the process of inverting data from other modalities to the language latent space in order to introduce new concepts in a pre-trained model~\cite{Gal:2022, Gal:2023, Daras:2022,wang2024conditional}. Gal~\etal~\cite{Gal:2022} formulate an optimisation task for image inversion with diffusion models, where visual reconstruction is used as guidance to find the token embedding that corresponds to the new concept. Daras~\etal~\cite{Daras:2022} extend this idea to enable multilevel resolution, with various levels of agreement to the given example concepts. This is achieved using the observation that the conditioning signal is dependent on the diffusion timestep. Instead of forming an optimization problem, Gal~\etal~\cite{Gal:2023} reformulate the problem as a regression based on the observation that one can underfit a large set of concepts from a given domain, and uses weight offsets to inject novel concepts of the same domain. In this work, we take inspiration from Gal~\etal~\cite{Gal:2022} and propose \emph{motion textual inversion}. We keep the initial formulation of using the reconstruction loss to learn the new embedding but apply it to a different realigned space. Our results suggest that this leads to improved realism and consistency with the example concepts.

% \paragraph{Diffusion.}
\section{Preliminaries}

Diffusion models are a class of generative models based on 

progressively corrupting data $\mathbf{x}_0$, where the noising process $\{ \mathbf{x}_t \}_{t=0}^{T}$ follows the Markovian principle: 
\begin{equation}
    q(\mathbf{x}_t | \mathbf{x}_{t-1}) = \mathcal{N}(\sqrt{
\alpha_{t}}\mathbf{x}_{t-1} , (1-\alpha_t)I),
\end{equation}
with $\alpha_t \in (0,1)$ hyperparameters that control the rate of diffusion at each timestep   . Note that when $\alpha_t$ is small, we can approximate $\mathbf{x}_t \sim \mathcal{N}(0, I)$. 
New samples can be generated by reversing the diffusion process, starting from a random vector $\mathbf{x}_T \sim \mathcal{N}(0, I)$ and predicting the next diffusion step iteratively. 
Here we follow the DDPM variant~\cite{Ho:2020} and train a neural network $\diffnet(\mathbf{x}_t, t)$ to predict the noise from step $t$ to $(t-1)$ by minimizing
\begin{equation}
    \mathbb{E}_{\mathbf{z}_0 \sim q(\mathbf{z}_0|c),t\sim[1,T]}||\epsilon -\diffnet(\mathbf{x}_t, t)||_2^2.
    \label{eq:simple_ldm}
\end{equation}
The model can additionally receive conditioning information $c$ such as an action category or language embedding, \ie $\epsilon_{t-1} = \diffnet(\mathbf{x}_t, t, c)$. 

An effective variant 
% gramatically incorect:
% was proven in the image domain 
is the Latent Diffusion Model~\cite{Roombach:2022, Chen:2023}, where diffusion is performed on the latent space of a pre-trained VAE instead of

\section{Method}

Our goal is to generate a motion sequence given a sentence in natural language. For this, we introduce ``Latent rEAlignment for human motion Diffusion'' (\modelname). We build on MLD~\cite{Chen:2023} and perform diffusion over the latent space of a previously trained motion VAE. Different from other approaches, \modelname includes a specialized \emph{projector} module trained to produce embeddings aligned with CLIP \cite{Radford:2021}, transforming the diffused latent into a better-structured semantic space (Section~\ref{sub:method}). Additionally, in Section~\ref{sub:mti} we introduce the task of \emph{motion textual inversion}, where we optimize for the textual embedding given a set of example motions. As will be shown in Section~\ref{Sec:experiments}, using the realigned space allows the optimization to better capture the input motion, showing the potential of our approach for personalized downstream tasks.

\begin{figure*}[h!]    
\includegraphics[width=\textwidth]{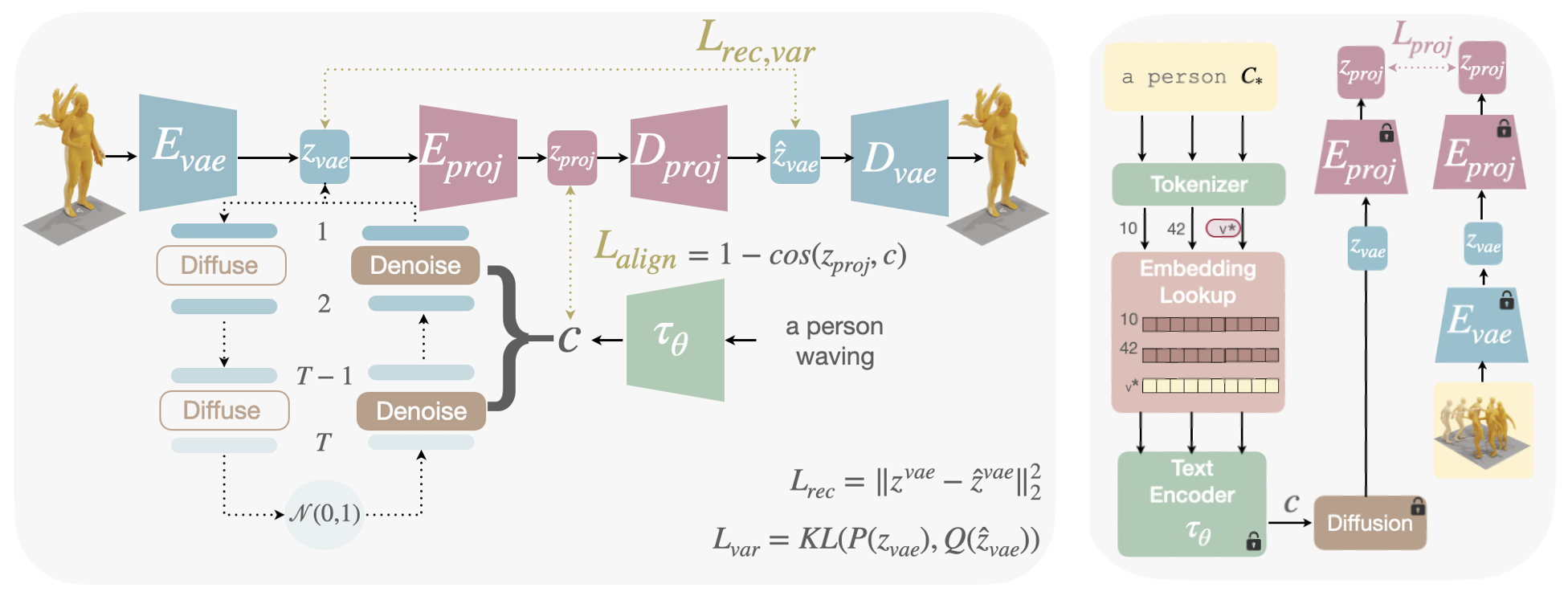}
    \caption{\textbf{\modelname}. 
    \textbf{Left: Text-to-motion generation with LEAD.}  
    \modelname consists of four modules: (1) A motion VAE (blue), a text encoder (green), a diffusion model (brown), and our new projector module (pink). Similar to latent diffusion model (LDM)~\cite{Roombach:2022}, we first train the VAE and then the diffusion model. We then train the projector module (pink) using an alignment loss towards the CLIP embedding, and a reconstruction loss towards the VAE embedding. 
    Once all modules are trained, we generate a motion latent $\vaeZ$ by sampling noise from the Gaussian distribution conditioned on the input text. The resulting latent is then auto-encoded by the projector and decoded through the VAE (blue) to obtain the final motion.
    \textbf{Right: Motion textual inversion.} 
    A pseudo-word ($C_*$) is added as an additional token, and we seek the optimal embedding $v_*$ to best reproduce the input. 
    Text conditioning guides the generation of motion through the diffusion module (brown). The embedding of the new token is learnt using the reconstruction objective on the realigned space.}
    \label{fig:method}
\end{figure*}

\subsection{\modelname}
\label{sub:method}

Given a sentence $y$, the goal of \modelname is to generate a motion sequence $\mathbf{x}\in \mathbb{R}^{N \times D}$, where $N$ is the motion length and $D$ the dimension of motion features, including joint rotations, positions, velocities and foot contacts as in ~\cite{Guo:2022}. 

\subsubsection{Architecture.} 
\modelname consists of four modules (Figure~\ref{fig:method}).

(1) A \textbf{motion VAE}~\cite{Kingma:2013:VAE} (blue in Fig.~\ref{fig:method}), with an encoder $\vaeZ_0 = \vaeE(\mathbf{x})$, where $\vaeZ_0  \in \mathbb{R} ^ M$ is a compressed representation of the motion segment, and a decoder $\tilde{\mathbf{x}} = \vaeD(\vaeZ_0)$ that transforms the latent back into a motion sequence. Following~\cite{Chen:2023}, the VAE is a transformer-based architecture~\cite{Vaswani:2017} with long skip connections~\cite{Ronneberger:2015:skipconnect}. The encoder $\vaeE$ takes as input two learnable distribution tokens corresponding to $\mu$ and $\sigma$ of a Gaussian distribution, along with the motion features. The VAE decoder $\vaeD$ takes as input the latent $\vaeZ$ and zero motion tokens and  generates the motion sequence via cross-attention. 

% \noindent
(2) A \textbf{conditional diffusion model} (brown in Fig.~\ref{fig:method})  
over the latent space, $\diffnet(\vaeZ_t, t, c)$, inspired by Motion-Latent-Diffusion (MLD)~\cite{Chen:2023}, that predicts noise given the current noised version of the latent $\vaeZ_t$, a timestep $t$, and a conditioning vector $c$ (text embedding). We employ here a transformer with long skip connections~\cite{Ronneberger:2015:skipconnect}. 

% \noindent
(3) A \textbf{text encoder} for the condition, $\tau_\theta$ (green in Fig.~\ref{fig:method}), which converts the input sentence into a latent embedding. 
% that is in turn used to guide the diffusion model $\diffnet$. 
In our work, $\tau_\theta$ follows the CLIP text-encoder~\cite{Radford:2021}.

% \noindent
(4) The \textbf{projector module} $\mathcal{P} = (\projE, \projD)$ (pink in Fig.~\ref{fig:method}), consisting of an encoder $\projE$ and a decoder $\projD$,  both of which are transformer-based architectures with sinusoidal positional encodings. 

\begin{figure*}[h!]
    \centering
    \includegraphics[width=1\linewidth]{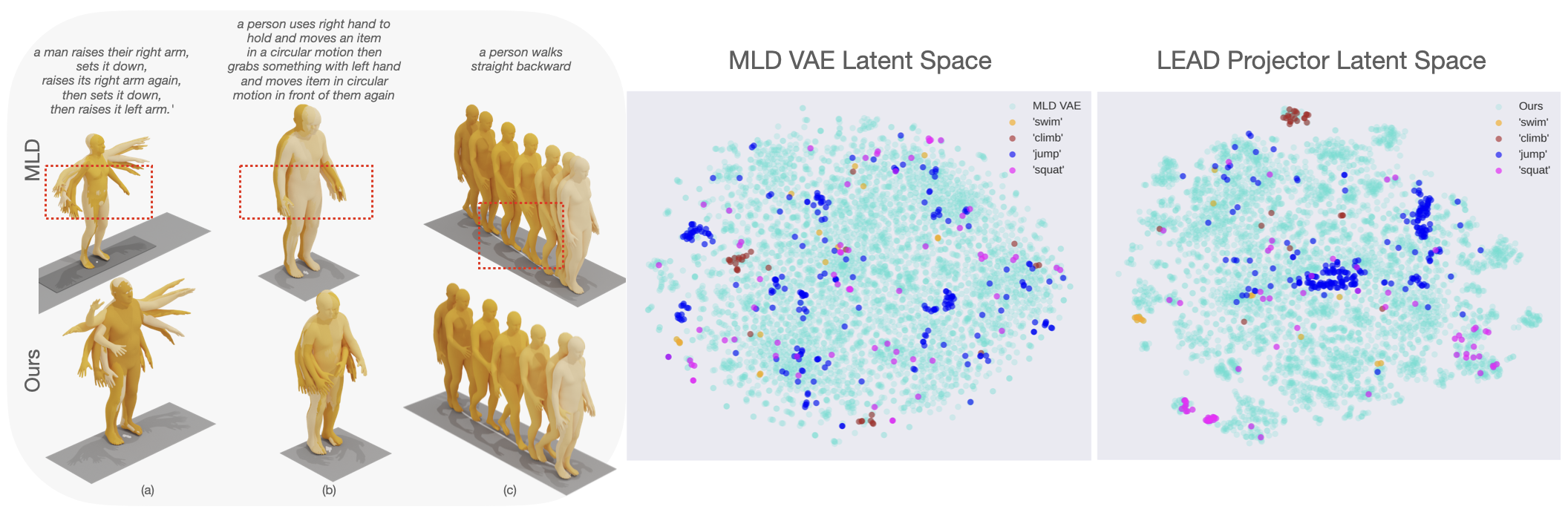}
    \caption{Left: Qualitative results for T2M compared to the baseline model MLD~\cite{Chen:2023}.
    Motions generated with our approach are more expressive and less static (a,b), and contain fewer artifacts like foot-sliding (c). 
    Right: Latent space visualization using tSNE~\cite{Maaten:2008:tSNE}. 
    }
    \label{fig:t2m}
\end{figure*}

The encoder transforms $\vaeZ$ into a new embedding $\projZ = \projE(\vaeZ)$, while the decoder transforms back into the VAE space, $\hat{\vaeZ} = \projD(\projZ)$ from which the motion can be recovered, $\hat{\mathbf{x}} = \vaeD(\hat{\vaeZ})$. Crucially, $\projE$ is trained to produce embeddings aligned with CLIP~\cite{Radford:2021}, such that they display a better semantic structure as shown in Fig.~\ref{fig:t2m}. 

 While text-to-image models are capable of producing high-quality samples, their performance is more limited in T2M, due to the smaller scale of datasets as well as the large gap between motion and language. The explicit realignment module aims to address these shortcomings by using multimodal cues, without requiring extra motion data.

\subsubsection{Training and losses.}  
We train \modelname in three stages. 
First, the motion VAE is trained over a large, unlabeled dataset consisting of motion data only, using the Mean Squared Error (MSE) and Kullback-Leibler divergence ($\mathop{KL}$) loss. \textcolor{red}{The MSE loss on the motion features acts as a geometric loss, ensuring that the motion latents retain geometric properties. }

Second, we freeze the VAE and the text-encoder $\tau_\theta$ and train the diffusion model $\diffnet$ using classifier-free guidance~\cite{Ho:2022}, which provides a trade-off between quality and diversity. This is done by applying $10\%$ dropout on the condition during training to learn both the conditioned and unconditioned distribution. 

Finally, we freeze the weights of $\diffnet$, $\tau_\theta$ and the VAE and train the realignment module $\mathcal{P}$, that is trained using multimodal information from motion and language. For this, we first project the motion sequence $\mathbf{x}$ into the VAE latent space, $\vaeZ=\vaeE(\mathbf{x})$. 
We next auto-encode $\vaeZ$ using $\mathcal{P}$:
\begin{equation}
\label{eq:proj_autoenc}
        \projZ = \projE(\vaeZ), \quad 
        \hat{\vaeZ} = \projD(\projZ).
\end{equation}

$\projE$ and $\projD$, are jointly trained using:

(a) An alignment loss $\mathcal{L}_{align}$ that measures the  similarity between the projected latent $z_{proj}$ and the corresponding CLIP text latent $c$ using cosine similarity:
 \begin{equation}
\mathcal{L}_{align}=1-\text{cos}(z_{proj},c).
    \label{eq:alignment}
\end{equation}

(b) A reconstruction loss $\mathcal{L}_{rec}$ that ensures we can recover the original VAE latent given the projected latent:

\begin{equation}
\mathcal{L}_{rec}= || \vaeZ - \hat{\vaeZ}||_2^2.
\end{equation}

(c) A variance loss $\mathcal{L}_{var}$ that ensures that the distribution of latents within a generated batch follows the distribution of latents within the training batch:
\begin{equation}
\mathcal{L}_{var}=\mathop{KL}(\mathop{P}(z_{vae}), \mathop{Q}(\hat{z}_{vae})),
\end{equation}
where $\mathop{P}, \mathop{Q}$ represent the probability distributions of ground-truth and predicted VAE latents.
As shown in Tab.~\ref{tab:ablations1}, this ensures that we retain the ability to generate diverse motions. 
The overall objective is:
\begin{equation}
    \mathcal{L} = \lambda_{align} \mathcal{L}_{align} + \lambda_{rec} \mathcal{L}_{rec} + \lambda_{var} \mathcal{L}_{var}.
    \label{eq:objective}
\end{equation}
where $\lambda_{align}$, $\lambda_{rec}$ and $\lambda_{var}$ weigh the corresponding terms.

\subsubsection{Inference.} 
% \noindent
% \textbf{Inference.} 
As shown in Fig.~\ref{fig:method}, following a standard reverse diffusion process, we sample a latent noise vector $\vaeZ_T \sim \mathcal{N}(0, I)$ and gradually denoise it using $\diffnet(\vaeZ_t, t, c)$ with the CLIP condition $c$ by relying on classifier-free guidance as in~\cite{Chen:2023}:%
\begin{equation}
\resizebox{0.9\columnwidth}{!}{
    $\diffnet^s(\vaeZ, t, c) = s (\diffnet (\vaeZ, t, c)) + (1-s)\diffnet (\vaeZ, t, \emptyset)$
}
\end{equation}
\noindent where $s$ denotes the guidance scale. 
Once obtained the clean latent $\vaeZ_0$, we pass it through the projector $\mathcal{P}$ and recover the output motion through 
$\mathbf{x} = \vaeD(\projD({\projZ}))$.

% --------------------------------------------------
\subsection{Motion Textual Inversion}
~\label{sub:mti}
\textcolor{red}{Latent motion diffusion models allow us to introduce} task of Motion Textual Inversion (MTI), where given a few examples of a motion, the goal is to find the corresponding embedding in the language space such that it can later be used to generate action sequences that retain the exemplar's characteristics. 

Following Gal~\etal~\cite{Gal:2022}, we assume that the concept to be learned can be captured using a single word ($C_*$).

Therefore, given a motion $m \in \mathbb{R}^{N\times D}$, we interpret the problem as seeking the optimal word embedding $v^*$ that best represents the concept in the latent space of a pre-trained text encoder, \ie CLIP. 
Similar to~\cite{Gal:2022}, the weights of the text encoder $\tau_{\theta}$, the VAE $(\vaeE, \vaeD)$ and the diffusion model $\diffnet$ are frozen, and only $v_*$ is modified. The motion concept depicted by the examples is represented through a new place-holder word $C_*$, along with its corresponding optimal $v_*$. The approach for motion textual inversion is illustrated in Figure~\ref{fig:method}.

While image textual inversion~\cite{Gal:2022} optimizes the MSE between reference and generated images, our initial experiments showed that a simple reconstruction loss (either on the motion features $\mathbf{x}$, or on the VAE latent $\vaeZ$) does not yield satisfactory results (see Sec.~\ref{sec:ablation}). We hypothesize that the proposed text-motion realignment mechanism can provide  more informative gradients during the token optimisation, leading to improve qualitative and quantitative performance. Based on our assumption, we propose minimizing the following loss function:

\begin{equation}
\label{eq: mti_z}
\big\| \projE(\mathop{F}(\epsilon)) - 
\projE(\mathop{F}(\epsilon_\theta(z_t, t, \tau_\theta(y'))) \big\|_2^2 
\end{equation}

\noindent
where $\mathop{F}$ denotes the conversion from a noised motion sample to a clean one with $\bar{\alpha}_t = \prod_{s=1}^t \alpha_s$, following~\cite{Ho:2020}:
\begin{equation}
x_0 \approx \hat{x}_0 = \mathop{F}(\epsilon) = (x_t - \sqrt{1-\bar{\alpha}_t \epsilon_\theta(x_t)}/\sqrt{\bar{\alpha}_t}
\label{eq: F}
\end{equation}

\noindent
\textbf{Template texts} When learning the pseudo-word of the new concept we use predefined template texts to guide the diffusion model. We define the template texts using examples from the HumanML3D dataset. Below we provide some examples of template texts: 
\begin{itemize}
    \item the sim $<$*$>$.
    \item the man/woman $<$*$>$.
    \item figure $<$*$>$.
    \item someone $<$*$>$.
    \item he/she $<$*$>$.
    \item  a robot $<$*$>$.
\end{itemize}

\section{Experiments}
\label{Sec:experiments}

In this section, we demonstrate the results of our method. 
First, we introduce the datasets and evaluation metrics (Sec.~\ref{sec:datasets}) as well as implementation details (Sec.~\ref{sec:implementation}). 
Next, we evaluate \modelname on the task of text-to-motion generation (T2M) (Sec.~\ref{sec:t2m_experiments}), and the newly proposed task of motion textual inversion (MTI) (Sec.~\ref{sec:mti_experiments}). 
Finally, in Sec.~\ref{sec:ablation} we ablate our design choices for both tasks. 
More qualitative results can be found in the supp. mat.

% -----------------------------------
\subsection{Datasets and Metrics}
\label{sec:datasets}

\subsubsection{Datasets.} 
For T2M we experiment on two standard datasets: HumanML3D~\cite{Guo:2022} and KIT-ML~\cite{Mandery:2015:KIT}. 
%Both datasets contain sequences of human poses accompanied by one or more textual descriptions. The \textbf{HumanML3D} dataset contains 14,616 motions and 44,970 descriptions, while \textbf{KIT-ML} contains 6,353 motions with 3,911 motion descriptions 
\textbf{HumanML3D} contains 14,616 human pose sequences with 44,970 descriptions, while \textbf{KIT-ML} contains 6,353 textual descriptions for 3,911 motions. Poses in both datasets are represented using the parameterization from~\cite{Guo:2022} consisting of the root angular velocity along the Y-axis, the root linear velocities on the XZ-plane, the root height, the root local joint positions, velocities and rotations in root space and binary features for foot-ground contact. 

For MTI, we perform all our quantitative comparisons on a randomly chosen subset of 100 motions from HumanML3D. This is due to the fact that we first need to optimize one placeholder token for each motion, which requires significant time. In addition, we qualitatively evaluate using the \textbf{100styles} dataset~\cite{Mason:2018}, which contains a wide range of locomotion in signature styles. \rev{In particular, this dataset contains out-of-distribution movements which are not used when training LEAD.} \rev{To accommodate the differences in the skeletal structure between 100styles dataset to our framework (HumanML3D),} we preprocess the dataset and obtain the representation as proposed in~\cite{Guo:2022}.

\noindent
\textbf{Motion Representation}
We adopt the HumanML3D representation~\cite{Guo:2022}, where each motion $\mathbf{x} \in \mathbb{R}^{N \times D}$ is expressed redundantly as $\mathbf{x} = \{\dot{r}_a, \dot{r}_x, \dot{r}_z, r_y,\mathbf{j}_p, \mathbf{j}_v, \mathbf{j}_r,\mathbf{c}_f \}$ that consists of:
\begin{itemize}
     \item  root angular velocity $\dot{r}_a \in \mathbb{R}$ along the Y-axis,
     \item  root linear velocity $\dot{r}_x, \dot{r}_z \in \mathbb{R}$ along the XZ-axis,
     \item root height $r_y \in \mathbb{R}$,
     \item local joint positions $\mathbf{j}_p \in \mathbb{R}^{3N_j}$,
     \item local joint velocities  $\mathbf{j}_v \in \mathbb{R}^{3N_j}$,
     \item local joint rotations  $\mathbf{j}_r \in \mathbb{R}^{6N_j}$ in the root space,
    \item binary foot contact labels $\mathbf{c}_f \in \mathbb{R}^{4}$ that are obtained by thresholding on the velocities of the heel and toe joints.
\end{itemize}
Here, $N_j$ denotes the number of joints ($N_J=22$ for the HumanML3D dataset and $N_j=21$ for KIT-ML).

\subsubsection{Evaluation Metrics.} 
% \noindent  \textbf{Evaluation Metrics.} 
For T2M we follow the standard evaluation protocol with metrics from~\cite{Guo:2020, Guo:2022} that 
measure four components:
motion realism, text-motion consistency, generation diversity and multimodal matching between text and motion. 
% 1
To assess realism we compute the Fréchet Inception Distance (\textbf{FID})~\cite{FID} between the ground-truth and predicted motion features, i.e. 
\begin{equation}
    \text{FID} = \mid \mid \mu_r - \mu_g \mid \mid^2 + Tr(\Sigma_r + \Sigma_g - 2(\Sigma_r\Sigma_g)^{\frac{1}{2}}),
\end{equation}
where $\mu_r,\Sigma_r$ and $\mu_g,\Sigma_g$ correspond to the mean and covariance matrix of the real and generated motions respectively.

\begin{figure*}[ht]
    \centering
     \begin{minipage}{0.24\textwidth}
  \begin{center}
    \resizebox{\textwidth}{!}{
    \includegraphics[width=3.5cm]{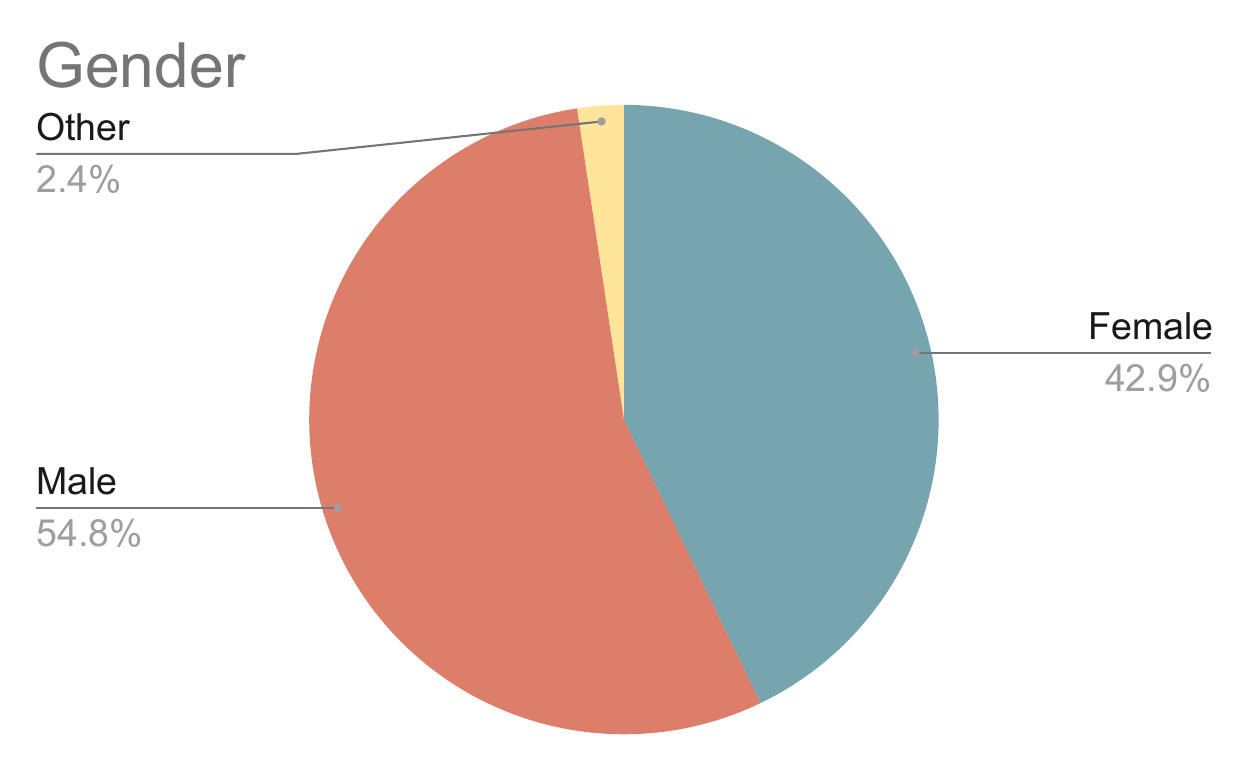} 
    }
    \captionof*{figure}{(a) Gender demographics.}
    \label{fig:stats1a}
    \end{center}
    \end{minipage}
    \hfill
    \begin{minipage}{0.24\textwidth}
  \begin{center}
    \resizebox{\textwidth}{!}{
    \includegraphics[width=3.5cm]{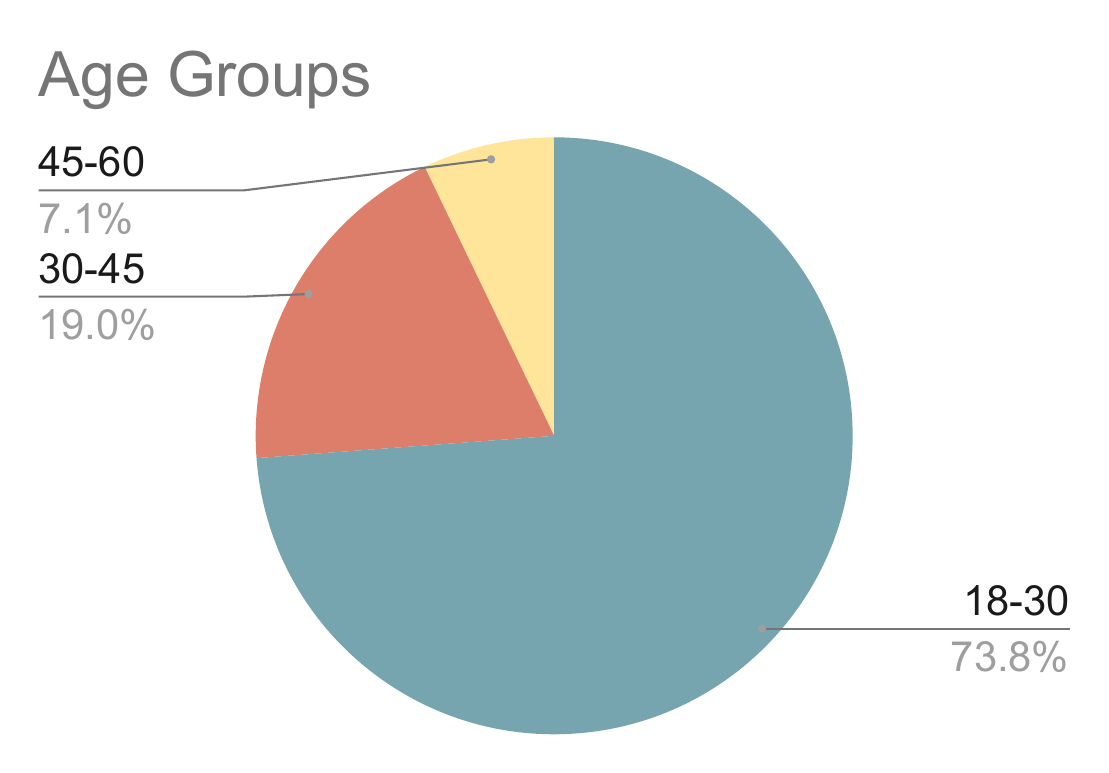} 
    }
    \captionof*{figure}{(b) Age demographics.}
    \label{fig:stats1b}
    \end{center}
    \end{minipage}
    \hfill
    \begin{minipage}{0.48\textwidth}
  \begin{center}
    \resizebox{0.8\textwidth}{!}{
    \includegraphics[width=1\linewidth]{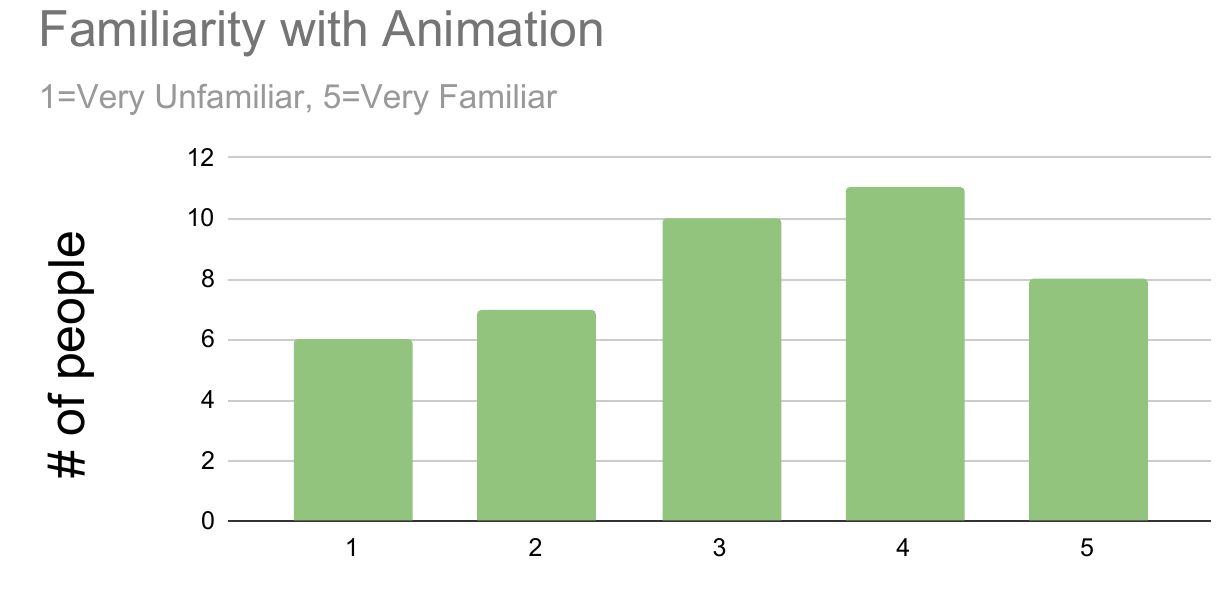}
    }
    \vspace{-11pt}
    \captionof*{figure}{(c) Familiarity of participants with computer animation.}
    \label{fig:stats2}
    \end{center}
    \end{minipage}
    \caption{User study information of participants.}
    \label{fig:user_study_stats}
\end{figure*}
    
   \begin{figure}[ht!]
    \centering
    % First: User study results (Realism and Text-Motion Relevance as subfigures)
    \subfloat[\centering Realism]{
        \includegraphics[width=3.5cm]{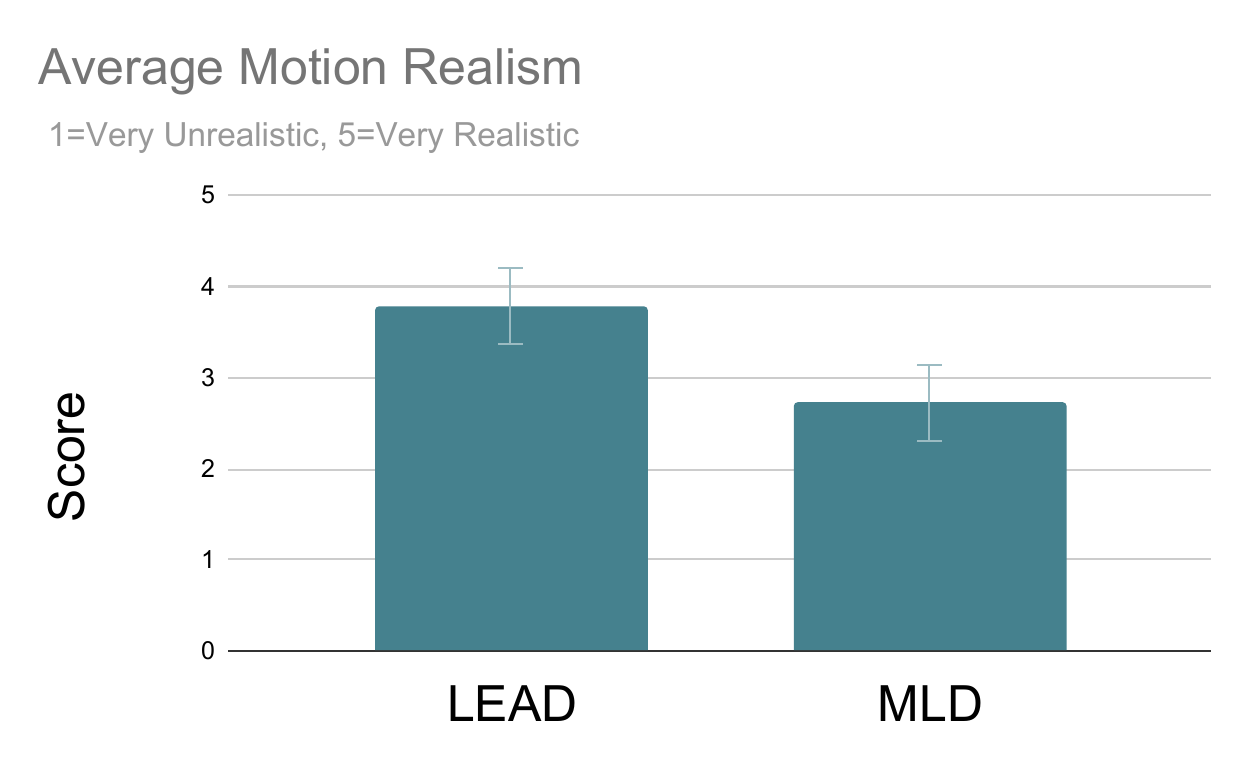}
    }
    \qquad
    \subfloat[\centering Text-Motion Relevance]{
        \includegraphics[width=3.5cm]{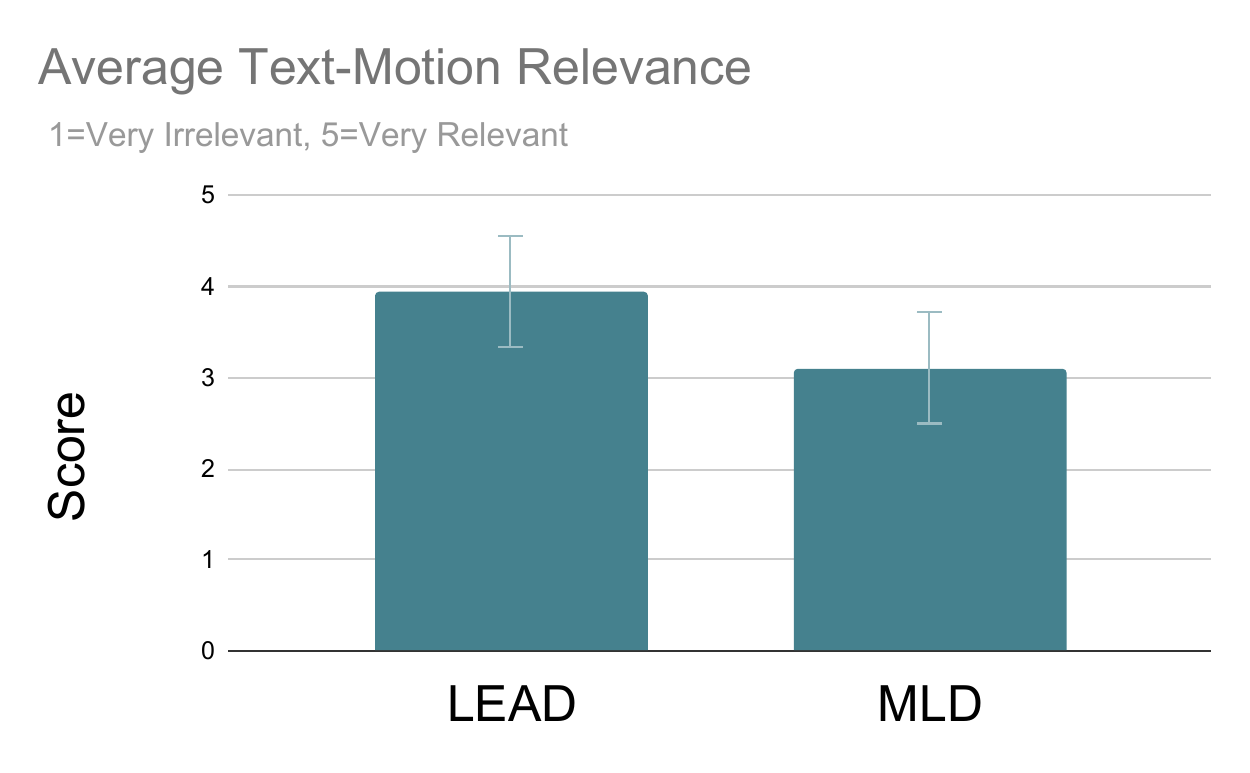}
    }
    
    % Second: Perceptual user study results (table)
    \vspace{1em} % Add some space between the figure and the table
    \subfloat[\centering \textcolor{red}{LEAD vs MotionCLIP user study results.}]{
\resizebox{0.45\textwidth}{!}{ % Scale to 80% of the text width
\begin{tabular}{c|c|c}
        & \textcolor{red}{Realism ($\uparrow$)} & \textcolor{red}{Text-Motion Relevance ($\uparrow$)} \\
        \midrule
        \midrule
        \textcolor{red}{MotionCLIP~\cite{Tevet:2022:MotionCLIP}} & \textcolor{red}{$1.9$} & \textcolor{red}{$1.6$}  \\
        \hline
        \textcolor{red}{LEAD} & \textcolor{red}{$\mathbf{4.0}$} & \textcolor{red}{$\mathbf{4.1}$} \\
        \bottomrule
\end{tabular}

}
  }

    % Common caption
    \caption{\textcolor{red}{User study results. We evaluate motion realism and text-motion relevance on a 1-5 scale where a higher score corresponds to better performance (1=very unrealistic/unrelated, 5=very realistic/related). Motions generated with \modelname are consistently perceived as more realistic and relevant than those generated using MLD~\cite{Chen:2023} (subfig (a) and (b)) and MotionCLIP~\cite{Tevet:2022:MotionCLIP} (subfig (c)).}}

    \label{fig:combined_results}
\end{figure}

% 2
To measure text-motion consistency we report the multimodal distance (MMdist) and R-precision, where MMdist measures the average Euclidean distance between the generated motions and the corresponding conditioning texts, and R-precision measures the average retrieval accuracy for the top 1/2/3 matches. We compute \textbf{MMdist} using 
\begin{equation}
    \text{MMdist} = \frac{1}{N} \sqrt{(\sum_{i=1}^{N} \|  v_i - t_i\|)}
\end{equation}
where $v_i$ are the features of the motions generated using texts $y_i$, and $t_i$ are the features of the corresponding texts, $y_i$. For \textbf{R-precision}, following Guo~\etal~\cite{Guo:2022}, for each generated motion we select its ground-truth textual description along with 31 randomly selected mismatched descriptions. Then we calculate the Euclidean distances between the motion feature and the 32 text features and rank them based on the distance. We select the top-1, top-2, and top-3 and consider as successful retrieval when the corresponding real text falls into the top-k candidates. R-precision is computed as the average across all generated motions from the test set.

% 3
To evaluate generation diversity (\textbf{Div}) we compute the variance of synthesized motions across all categories. For a subset of size $p$ the diversity can be computed as:
\begin{equation}
    \text{DIV} = \frac{1}{p}\sum_{i=1}^p\mid\mid v_i - \hat{v}_i \mid\mid
\end{equation}
where $v_i$, and $\hat{v}_i$ are the features corresponding to the first and second subset of predictions respectively. Following previous work~\cite{Chen:2023}, we set $p=300$. 

% 4
Finally, multimodality (MModality) measures the variance across 100 motions generated using the same textual description. To calculate \textbf{MModality} (multimodality) we sample $m$ textual descriptions. For each description we generate two subsets of motions of size d. Then MModality is computed as:
\begin{equation}
    \text{MModality} = \frac{1}{m \times d}\sum_{j=1}^m\sum_{i=1}^d\mid\mid v_{j,i} - \hat{v}_{j,i} \mid \mid
\end{equation}
where $v$ and $\hat{v}$ are the motion features of each subset. Following~\cite{Chen:2023} we set $m=100$ and $d=10$. 
 
The metrics are computed on the features extracted by a pre-trained motion feature extractor~\cite{Guo:2022}.

% -----------------------------------
\subsection{Implementation details}
\label{sec:implementation}

For the text encoder we employ CLIP-ViT-14~\cite{Radford:2021}.
The projector module is trained for 200 epochs with AdamW optimizer, which lasts approximately 2 hours on an NVIDIA GeForce 3090. While training the realignment module we keep the text encoder and the VAE frozen. 
We use a learning rate of $1e^{-4}$, batch size $1024$ with optimal hyperparameters $\lambda_{rec}=4.7036$, $\lambda_{align}=2.1591$ and $\lambda_{var}=0.05960$ (Eq.~\ref{eq:objective}).

For T2M, we retain the hyperparameters from MLD~\cite{Chen:2023}, with 1K diffusion steps and 50 for training and inference respectively. For MTI, the pseudo-word embedding is initialized using a coarse one-word descriptor. We optimise for 20 steps using ADAM optimizer and batch of size 4, consisting of motion segments derived from one motion sequence.

The projector is realised as a transformer-based encoder/decoder. The encoder and decoder both consist of 9 layers with 4 heads and hidden size 1024. We use 0.1 dropout and the GeLu activation function~\cite{Hendrycks:2016:GELU}. We use sinusoidal positional encodings.
\begin{table*}[ht!]
\centering
\small
\caption{T2M on HumanML3D. Numbers are taken from previous works. Bold and underline denote first and second best, respectively. \textcolor{red}{The second column indicates which models follow the latent diffusion (LDM) architecture. LEAD is designed for latent diffusion models (LDM), as they enable the downstream application of MTI. By incorporating LEAD, performance is further improved in LDM models.}} 
\label{tab:metrics1}
\resizebox{\linewidth}{!}{%
\begin{tabular}{l||cccccccc}
\toprule
\multirow{1}{*}{} & \multicolumn{1}{c}{\parbox{20pt}{LDM}} & \multicolumn{1}{c}{FID ($\downarrow$)}  & \multicolumn{3}{c}{R-precision ($\uparrow$)} & \multicolumn{1}{c}{MMdist ($\downarrow$)} & \multicolumn{1}{c}{Diversity ($\rightarrow$)} & \multicolumn{1}{c}{MModality ($\uparrow$)}  \\
\cmidrule(lr){2-2}  \cmidrule(lr){3-3}  \cmidrule(lr){4-6} \cmidrule(lr){7-7} \cmidrule(lr){8-8}  \cmidrule(lr){9-9}
& & & top1 & top2 & top3 & & & \\
\midrule
\midrule
Real & & $0.002^{\pm .008}$ & $0.511^{\pm .003}$ & $0.703^{\pm .003}$ & $0.797^{\pm .002}$ & $2.974^{\pm .008}$ & $9.503^{\pm .065}$ & -  \\
\hline
JL2P~\cite{Ahuja:2019}& & $11.02^{\pm .046}$  & $0.246^{\pm .001}$ & $0.387^{\pm .002}$ & $0.486^{\pm .002}$ & $5.296^{\pm .008}$ & $7.676^{\pm .058}$ & - \\
T2G~\cite{Bhattacharya:2021} & & $7.664^{\pm .030}$  & $0.165^{\pm .001}$ & $0.267^{\pm .002}$ & $0.345^{\pm .002}$ & $6.030^{\pm .008}$ & $6.409^{\pm .071}$ & -\\
TEMOS~\cite{Petrovich:2022} & & $3.734^{\pm .028}$  & $0.424^{\pm .002}$ & $0.612^{\pm .002}$ & $0.722^{\pm .002}$ & $3.703^{\pm .008}$ & $8.973^{\pm .071}$ & $0.368^{\pm .018}$ \\
T2M~\cite{Guo:2022}& & $1.067^{\pm .002}$  & $0.457^{\pm .002}$ & $0.639^{\pm .003}$ & $0.740^{\pm .003}$ & $3.340^{\pm .008}$ & $9.188^{\pm .002}$ & $2.090^{\pm .083}$ \\
MDM~\cite{Tevet:2022}& & $0.544^{\pm .044}$  & $0.320^{\pm .005}$ & $0.498^{\pm .004}$ & $0.611^{\pm .007}$ & $5.566^{\pm .027}$ & ${9.559^{\pm .086}}$ & $\secondbest{2.799^{\pm0.072}}$ \\
MotionDiffuse~\cite{Zhang:2022}& &$0.630^{\pm .001}$ & $0.491^{\pm .001}$ & $0.681^{\pm .001}$ & {$0.782^{\pm .001}$} & {$3.113^{\pm .001}$} & {$9.410^{\pm .049}$} & $1.553^{\pm .042}$ \\
TM2T~\cite{Guo:2022:tm2t}& & $1.501^{\pm .017}$  & $0.424^{\pm .003}$ & $0.618^{\pm .001}$ & $0.729^{\pm .002}$ & $3.467^{\pm .011}$ & $8.589^{\pm .076}$ & $2.424^{\pm .093}$ \\
T2M-GPT~\cite{Zhang:2023} & &$ 0.116^{\pm .004}$  & ${0.491^{\pm .003}}$ & ${0.680^{\pm .003}}$ & ${0.775^{\pm .002}}$ & ${3.118^{\pm .011}}$ & $9.761^{\pm .081}$ & $1.856^{\pm .011}$\\
GMD~\cite{Karunratanakul:2023}& & $ 0.212^{\pm .000}$ & - & - &  $0.670^{\pm .000}$ & - &  ${9.440^{\pm .000}}$ & - \\
Fg-T2M~\cite{Wang:2023}& & $ 0.243^{\pm .019 }$  & $ {0.492^{\pm .002 }}$ &  ${0.683^{\pm .003 }}$ &  ${0.783^{\pm .002}}$ &${3.109^{\pm .007}}$&  $9.278^{\pm .072}$ & $1.614^{\pm .049}$ \\
ReMoDiffuse~\cite{Zhang:2023:ReMoDiffuse} & & $\best{0.103^{\pm .004}}$ & $\secondbest{0.510^{\pm .005}}$ & $\secondbest{0.698^{\pm .006}}$& $\secondbest{0.795^{\pm .004}}$ & $\best{2.794^{\pm .016}}$ & $9.018^{\pm .075}$ & $1.796^{\pm .043}$ \\
OmniControl~\cite{xie2023omnicontrol} & & $0.310^{\; \quad \quad}$ &  $-$ & $-$& $0.693^{\; \quad \quad}$ & $-$ & $\best{9.502^{\; \quad \quad}}$ & $-$ \\
MotionGPT~\cite{jiang2024motiongpt} & & $0.232^{\pm .008}$  & ${0.492^{\pm .003}}$ & $0.681^{\pm .003}$& $0.778^{\pm .002}$ & ${3.096^{\pm .008}}$ & $\secondbest{9.528^{\pm .071}}$ & ${2.008^{\pm .084}}$ \\ 
\hline
MotionLCM~\cite{motionlcm} & \checkmark & ${{0.467} ^{\pm .012}}$  & $0.502 ^{\pm .003}$ & $0.701 ^{\pm .002}$ & $0.803 ^{\pm .002}$ & $3.022^{\pm .009}$ & ${9.631 ^{\pm .066}}$ & ${2.172 ^{\pm .082}}$ \\
\textbf{\modelname (MotionLCM) (Ours)}
& \checkmark & $0.296^{\pm .007}$ & $\best{0.522^{\pm .003}}$ & $\best{0.721}^{\pm .003}$ & $\best{0.818^{\pm .003}}$ & $\secondbest{2.870^{\pm .010}}$ & $9.820^{\pm .074}$ & $1.974^{\pm .054}$ \\
\hline
MLD~\cite{Chen:2023}& \checkmark &$0.473^{\pm .013}$  & ${0.481^{\pm .003}}$ & $0.673^{\pm .003}$ & $0.772^{\pm .002}$ & $3.196^{\pm .010}$ & $9.724^{\pm .082}$ & $2.413^{\pm .079}$  \\
\textbf{\modelname (MLD) (Ours)} & \checkmark & ${\secondbest{0.109} ^{\pm .005}}$  & $0.464 ^{\pm .003}$ & $0.649 ^{\pm .004}$ & $0.743 ^{\pm .003}$ & $3.324^{\pm .015}$ & ${9.627 ^{\pm .146}}$ & $\best{2.902 ^{\pm .015}}$  \\
\bottomrule
\end{tabular}}
\end{table*}

\begin{table*}[h!]
\centering
\caption{T2M on the KIT-ML dataset. Numbers are taken from previous works. Bold and underline denote first and second best, respectively. \textcolor{red}{The second column indicates which models follow the latent diffusion (LDM) architecture. LEAD is designed for latent diffusion models (LDM), as they enable the downstream application of MTI. By incorporating LEAD, performance is further improved in LDM models.}}
\label{tab:metrics2}
\resizebox{\linewidth}{!}{%
\begin{tabular}{l||cccccccc}
\toprule
\multirow{1}{*}{} & \multicolumn{1}{c}{\parbox{20pt}{LDM}} & \multicolumn{1}{c}{FID ($\downarrow$)}  & \multicolumn{3}{c}{R-precision ($\uparrow$)} & \multicolumn{1}{c}{MMdist ($\downarrow$)} & \multicolumn{1}{c}{Diversity ($\rightarrow$)} & \multicolumn{1}{c}{MModality ($\uparrow$)}  \\
\cmidrule(lr){2-2}  \cmidrule(lr){3-3}  \cmidrule(lr){4-6} \cmidrule(lr){7-7} \cmidrule(lr){8-8}  \cmidrule(lr){9-9}
& & & top1 & top2 & top3 & & & \\
\midrule
\midrule
Real & & $0.031^{\pm .004}$  & $0.424^{\pm .005}$ & $0.649^{\pm .006}$ & $0.779^{\pm .006}$ & $2.788^{\pm .012}$ & $11.08^{\pm .097}$ & - \\
\hline 
LJ2P~\cite{Ahuja:2019} & & $6.545^{\pm .072}$ &  $0.221^{\pm .005}$ & $0.373^{\pm .004}$ & $0.483^{\pm .005}$ & $5.147^{\pm .030}$ & $9.073^{\pm .100}$ & - \\
T2G~\cite{Bhattacharya:2021}& & $12.12^{\pm .183}$ &  $0.156^{\pm .004}$ & $0.255^{\pm .004}$ & $0.338^{\pm .005}$ & $6.964^{\pm .029}$ & $9.334^{\pm .079}$ & - \\
TEMOS~\cite{Petrovich:2022} & & $3.717^{\pm .051}$ &  $0.353^{\pm .006}$ & $0.561^{\pm .007}$ & $0.687^{\pm .005}$ & $3.417^{\pm .019}$ & $10.84^{\pm .100}$ & $0.532^{\pm .034}$ \\
T2M ~\cite{Guo:2022}& & $2.770^{\pm .109}$ & $0.370^{\pm .005}$ & $0.569^{\pm .007}$ & $0.693^{\pm .007}$ & $3.401^{\pm .008}$ & $10.91^{\pm .119}$ & $1.482^{\pm .065}$ \\
MDM~\cite{Tevet:2022}& & $0.497^{\pm .021}$ &  $0.164^{\pm .004}$ & $0.291^{\pm .004}$ & $0.396^{\pm .004}$ & $9.191^{\pm .022}$ & $10.85^{\pm .109}$ & $1.907^{\pm .214}$ \\
MotionDiffuse~\cite{Zhang:2022}& & $1.954^{\pm .062}$ &  {$0.417^{\pm .004}$} & {$0.621^{\pm .004}$} & {$0.739^{\pm .004}$ }& $\best{2.958^{\pm .005}}$ & $\secondbest{11.10^{\pm .143}}$ & $0.730^{\pm .013}$ \\
TM2T\cite{Guo:2022:tm2t}& & $3.599^{\pm .153}$ & $0.280^{\pm .005}$ & $0.463^{\pm .006}$ & $0.587^{\pm .005}$ & $4.591^{\pm .026}$ & $9.473^{\pm .117}$ & $\best{3.292^{\pm .081}}$ \\
T2M-GPT~\cite{Zhang:2023} & & $0.514^{\pm .029}$ &  ${0.416}^{\pm .006}$ & $\secondbest{{0.627}^{\pm .006}}$ & $\secondbest{0.745^{\pm .006}}$ &{$3.007^{\pm .023}$} & $10.921^{\pm .108}$ & $1.570^{\pm .039}$ \\
Fg-T2M~\cite{Wang:2023}& & $ 0.571^{\pm .047 }$ &  $\secondbest{ 0.418^{\pm .005 }}$ & $ 0.626^{\pm .004 }$ &  $ \secondbest{0.745^{\pm .004}}$ &$3.114^{\pm .015}$&  $10.93^{\pm .083}$ & $1.019^{\pm .029}$\\
ReMoDiffuse~\cite{Zhang:2023:ReMoDiffuse} & & $\best{0.155^{\pm .006}}$ & $\best{0.427^{\pm .014}}$ & $\best{0.641^{\pm .004}}$& $\best{0.765^{\pm .055}}$ & $\secondbest{2.814^{\pm .012}}$ & $10.80^{\pm .105}$ & $1.239^{\pm .028}$ \\
OmniControl~\cite{xie2023omnicontrol} & & $0.788^{\; \quad \quad}$  & $-$ & $-$& $0.379^{\; \quad \quad}$ & $-$ & ${10.841^{\; \quad \quad}}$ & $-$ \\
MotionGPT~\cite{jiang2024motiongpt} & & $0.510^{\pm .016}$ & ${0.366^{\pm .005}}$ & $0.558^{\pm .004}$& $0.680^{\pm .005}$ & ${3.527^{\pm .021}}$ & ${10.350^{\pm .084}}$ & $\secondbest{2.328^{\pm .117}}$ \\
\hline
MLD~\cite{Chen:2023}& \checkmark & $0.404^{\pm .027}$  & $0.390^{\pm .008}$ & $0.609^{\pm .008}$ & $0.734^{\pm .007}$ & $3.204^{\pm .027}$ & $10.800^{\pm .117}$ & {$2.192^{\pm .071}$}  \\
\textbf{\modelname (MLD) (Ours)} & \checkmark & $\secondbest{{0.246}^{\pm .014}}$ & $0.388^{\pm .005}$ & $0.608^{\pm .005}$ & $0.732^{\pm .005}$ & $3.199^{\pm .031}$ & $\best{{11.099}^{\pm .080}}$ & $1.919^{\pm .060}$ \\
\bottomrule
\end{tabular}}
\end{table*}

\subsection{Text-to-Motion Generation}
\label{sec:t2m_experiments}

For T2M, we compare against the SOTA on the HumanML3D and KIT-ML datasets. We report the average statistics and 95\% confidence interval across 20 runs.

Tab.~\ref{tab:metrics1} report the results for HumanML3D and Tab.~\ref{tab:metrics2} for KIT-ML. Notably, on both datasets, \modelname achieves a significant improvement in motion realism compared to the other methods, as reflected by the FID. This includes the reference \textcolor{red}{latent diffusion methods (MLD~\cite{Tevet:2022}, MotionLCM~\cite{motionlcm})} which differ only in the use of the projector, showing that a simple realignment step during inference can substantially boost realism. 
Overall, we observe that the realignment mechanism leads to significant improvements in motion realism over the baselines (MLD, MotionLCM). Even though it demonstrates slightly worse performance compared to the state-of-the-art ReMoDiffuse, which relies on a retrieval mechanism at inference, our method achieves a better trade-off between realism and alignment as reflected in SOTA performance (R-precision, MMdist).

Additionally, we visualize using tSNE~\cite{Maaten:2008:tSNE} the VAE latent space in comparison to the realigned space. As seen in Fig.~\ref{fig:t2m} (right), similar motions are better clustered in the projected space, 

suggesting that a semantic structure of the latent space might have an effect on the generation quality.

Finally, we show qualitative results in Fig.~\ref{fig:t2m}. We observe that the synthesized motions from \modelname are more lifelike and expressive compared to the ones from MLD. In particular, Fig.~\ref{fig:t2m} (a,b) show that motions generated with \modelname display richer movement that still complies with the text, while motions generated with MLD are smoother and inert, an observation that holds for more rare actions such as swimming. Furthermore, \modelname minimizes motion artefacts such as foot-sliding, as can be seen in Fig.~\ref{fig:t2m} (c). We provide animated qualitative results and comparisons to baselines in the \textbf{supplementary video}. We compare text-to-motion results of \modelname with those generated using T2M~\cite{Guo:2022}, MDM~\cite{Tevet:2022} and our reference method, MLD~\cite{Chen:2023}.

\begin{table}[h]
\caption{MTI evaluation on mini-subset of HumanML3D. We show the results using the reconstruction objective on the VAE space of MLD ($\text{MTI}_{mld}$)}, the motion feature space ($\text{MTI}_{feat}$) and our realigned space ($\text{MTI}_{lead}$), with (w/project) and without autoencoding through the projector during generation.
\label{tab:mti}
\resizebox{\linewidth}{!}{%
\begin{tabular}{l|c||HHcccc}
\toprule
\multirow{2}{*}{ } & Project & \multicolumn{3}{c}{R-prec. (top3) ($\uparrow$)} & \multicolumn{1}{c}{FID ($\downarrow$)} & \multicolumn{1}{c}{MMdist ($\downarrow$)} & \multicolumn{1}{c}{Div.($\rightarrow$)}  \\

\midrule
\midrule
Real & -- & 
0.511\rmstd{\textsuperscript{$\pm$0.023}} &
0.712\rmstd{\textsuperscript{$\pm$0.014}}
& 0.795\rmstd{\textsuperscript{$\pm$0.011}} &- & 2.967\rmstd{\textsuperscript{$\pm$0.079}} & 9.681\rmstd{\textsuperscript{$\pm$0.279}} \\
\hline

$\text{MTI}_{feat}$ & -- &  
0.233\rmstd{\textsuperscript{$\pm$0.020}}&
0.358\rmstd{\textsuperscript{$\pm$0.016}} &
0.447\rmstd{\textsuperscript{$\pm$0.012}} &
5.929\rmstd{\textsuperscript{$\pm$0.256}} &
5.360\rmstd{\textsuperscript{$\pm$0.065}} &
8.296\rmstd{\textsuperscript{$\pm$0.278}} \\
$\text{MTI}_{feat}$ & \checkmark & 
0.244\rmstd{\textsuperscript{$\pm$0.016}} &
0.377\rmstd{\textsuperscript{$\pm$0.015}} &
0.473\rmstd{\textsuperscript{$\pm$0.016}} &
4.754\rmstd{\textsuperscript{$\pm$0.290}} &
5.244\rmstd{\textsuperscript{$\pm$0.064}} &
8.396\rmstd{\textsuperscript{$\pm$0.257}}\\
\hline
$\text{MTI}_{mld}$ & -- & 
0.287\rmstd{\textsuperscript{$\pm$0.022}} & 0.432\rmstd{\textsuperscript{$\pm$0.023}} & 0.528\rmstd{\textsuperscript{$\pm$0.021}} & 3.066\rmstd{\textsuperscript{$\pm$0.172}} & 4.741\rmstd{\textsuperscript{$\pm$0.114}} & \textbf{8.874\rmstd{\textsuperscript{$\pm$0.178}}} \\
$\text{MTI}_{mld}$ &\checkmark & 
0.298\rmstd{\textsuperscript{$\pm$0.020}} & 0.447\rmstd{\textsuperscript{$\pm$0.023}} & \textbf{0.537\rmstd{\textsuperscript{$\pm$0.021}}} & 2.736\rmstd{\textsuperscript{$\pm$0.174}} & 4.679\rmstd{\textsuperscript{$\pm$0.116}} & 8.768\rmstd{\textsuperscript{$\pm$0.187}} \\
\hline
$\text{MTI}_{lead}$ & -- & 
0.285\rmstd{\textsuperscript{$\pm$0.016}} & 0.431\rmstd{\textsuperscript{$\pm$0.017}} & 0.521\rmstd{\textsuperscript{$\pm$0.015}} & 3.042\rmstd{\textsuperscript{$\pm$0.197}} & 4.760\rmstd{\textsuperscript{$\pm$0.074}} & 8.741\rmstd{\textsuperscript{$\pm$0.193}} \\

$\text{MTI}_{lead}$ &  \checkmark& 
0.296\rmstd{\textsuperscript{$\pm$0.015}} & 0.437\rmstd{\textsuperscript{$\pm$0.014}} & 0.533\rmstd{\textsuperscript{$\pm$0.013}} & \textbf{2.529\rmstd{\textsuperscript{$\pm$0.178}}} & \textbf{4.663\rmstd{\textsuperscript{$\pm$0.064}}} & 8.741\rmstd{\textsuperscript{$\pm$0.207}} \\
\bottomrule
\end{tabular}}
\end{table}

\begin{table}[t]
\caption{MTI ablation on mini-subset of HumanML3D. We report results with the realignment scheme during generation.}

\label{tab:ablations2}
\resizebox{\linewidth}{!}{%
\begin{tabular}{l|c||HHcccc}
\toprule
\multirow{2}{*}{ } & Project & \multicolumn{3}{c}{R-prec. (top3) ($\uparrow$)} & \multicolumn{1}{c}{FID ($\downarrow$)} & \multicolumn{1}{c}{MMdist ($\downarrow$)} & \multicolumn{1}{c}{Div.($\rightarrow$)}  \\

\midrule
\midrule

Real & -- & 
0.511\textsuperscript{$\pm$0.023} &
0.712\textsuperscript{$\pm$0.014}
& 0.795\rmstd{\textsuperscript{$\pm$0.011}} &- & 2.967\rmstd{\textsuperscript{$\pm$0.079}} & 9.681\rmstd{\textsuperscript{$\pm$0.279}} \\
\hline

noALIGN & \checkmark &
0.304\rmstd{\textsuperscript{$\pm$0.015}} &
0.555\rmstd{\textsuperscript{$\pm$0.016}} &
\textbf{0.539\rmstd{\textsuperscript{$\pm$0.018}} }&
2.762\rmstd{\textsuperscript{$\pm$0.204}} &
\textbf{4.642\rmstd{\textsuperscript{$\pm$0.071}}} &
8.729\rmstd{\textsuperscript{$\pm$0.197}} \\

noKL  & \checkmark &
0.289\rmstd{\textsuperscript{$\pm$0.011}} & 0.438\rmstd{\textsuperscript{$\pm$0.015}} & 0.535\rmstd{\textsuperscript{$\pm$0.017}} & 2.963\rmstd{\textsuperscript{$\pm$0.226}} & 4.710\rmstd{\textsuperscript{$\pm$0.075}} & \textbf{8.945\rmstd{\textsuperscript{$\pm$0.189}}} \\

noREC  & \checkmark &
0.296\rmstd{\textsuperscript{$\pm$0.017}} & 0.435\rmstd{\textsuperscript{$\pm$0.017}} & 0.526\rmstd{\textsuperscript{$\pm$0.015}} & 3.537\rmstd{\textsuperscript{$\pm$0.196}} & 4.725\rmstd{\textsuperscript{$\pm$0.079}} & 8.373\rmstd{\textsuperscript{$\pm$0.205}} \\

PROJ & \checkmark &
0.296\rmstd{\textsuperscript{$\pm$0.015}} & 0.437\rmstd{\textsuperscript{$\pm$0.014}} & 0.533\rmstd{\textsuperscript{$\pm$0.013}} & \textbf{2.529\rmstd{\textsuperscript{$\pm$0.178}}} & 4.663\rmstd{\textsuperscript{$\pm$0.064}} & 8.741\rmstd{\textsuperscript{$\pm$0.207}}\\
\bottomrule
\end{tabular}}
\end{table}

\noindent
\textbf{User Study} To assess the quality of our generated motions we conducted two user studies, where we compared LEAD against our reference methods MLD and MotionCLIP on two axes: motion realism and text-motion relevance. 

For this, we visualized 10 pairs of motions side-by-side for each method and asked participants to rate on a scale of 1-5 (1=very unrealistic/unrelated, 5=very realistic/related). \textcolor{red}{We report our findings in Fig.~\ref{fig:combined_results}}.

\textbf{LEAD vs MLD:} We first assessed motion realism (ignoring the text), where \modelname achieved 3.81 on average while MLD achieved 2.75, showing a significant outperformance %wrt realism 
when applying the projector module. Secondly, we showed the text description used to generate the motions and asked the users to rate how well the text and motion were matched. Here, \modelname scores 3.97 while MLD scores 3.12, showing that our generated motions are not only realistic but can also better match the input text. More details of the user study can be found in the supplementary. For the user study, we select motions generated using \modelname and MLD, conditioned on textual descriptions from HumanML3D~\cite{Guo:2022}. Our selection is based on the criterion of how much the motion differs before and after the realignment. We enlisted 40 participants with various age, gender and knowledge backgrounds (see Fig.~\ref{fig:user_study_stats}). Our results in terms of average motion realism and text-motion relevance can be seen in Fig.~\ref{fig:combined_results}, and clearly demonstrate a preference for our method.

\begin{figure}
    \centering
    \includegraphics[width=1\linewidth]{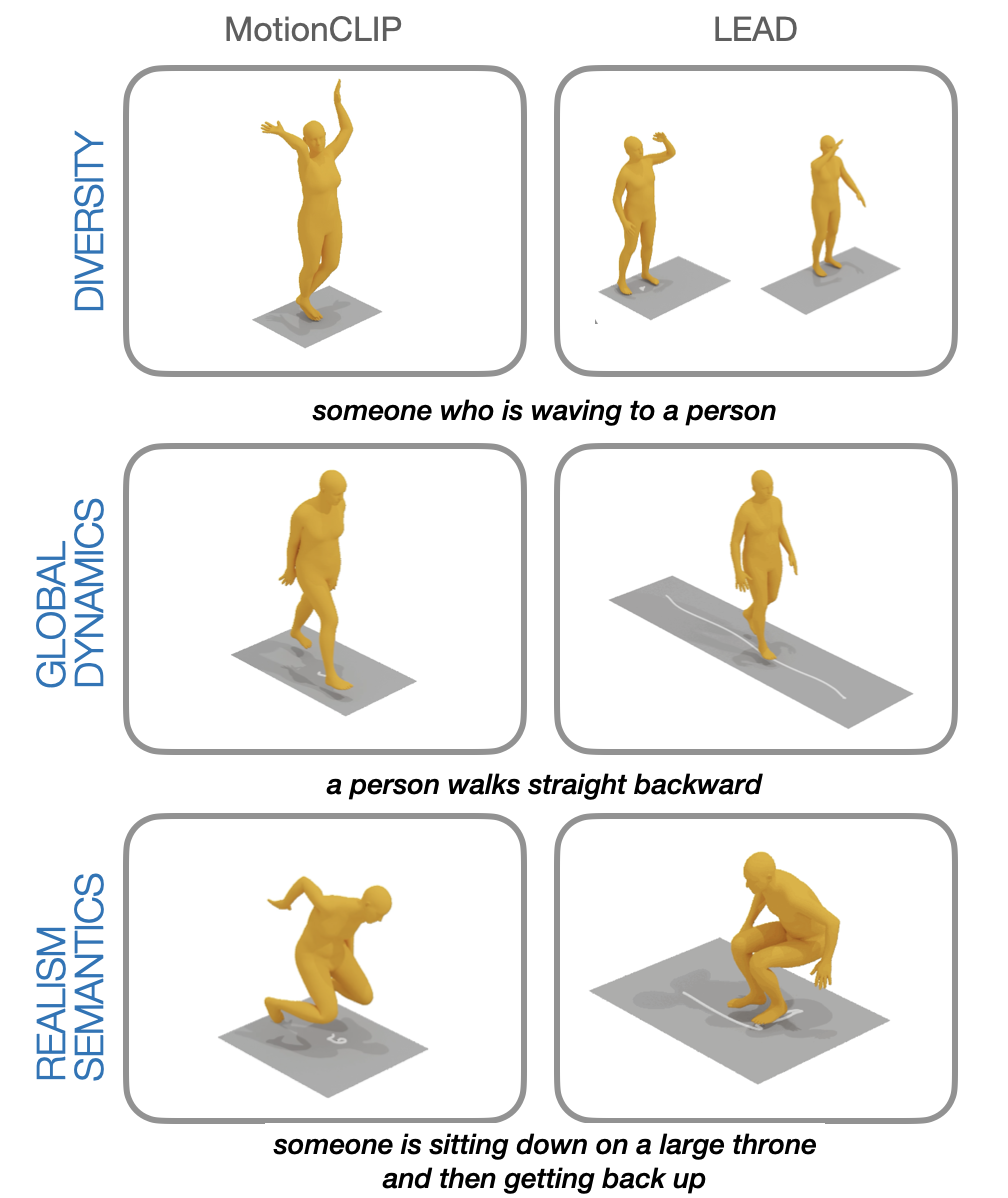}
    \caption{Qualitative comparison between LEAD and MotionCLIP on 4 axes: diversity, global dynamics, realism and semantic alignment.}
    \label{fig:LEADvsMotionCLIP}
\end{figure}

\color{red}{\textbf{LEAD vs MotionCLIP:}{
Fig.~\ref{fig:combined_results} shows the average results in a user study of 30 participants. On motion realism (ignoring the text), \modelname significantly outperforms MotionCLIP achieving average scores of 4.0 and 1.9 respectively. In terms of text-motion alignment, \modelname scores 4.1 while MotionCLIP scores 1.6, showing that our generated motions are not only realistic but can also better match the input text. Qualitative results as seen in Fig.~\ref{fig:LEADvsMotionCLIP} show that LEAD outperforms MotionCLIP in realism and semantic alignment, while additionally capturing motion diversity and generating global motion dynamics (translation, orientation). }}
\color{black}{}

\noindent
\textbf{Inference Time}
We provide a comparison in terms of inference times between \modelname and MLD~\cite{Chen:2023}. We find that the performance benefits of \modelname come with marginally increased computational overhead compared to MLD~\cite{Chen:2023}. Inference takes 0.236s per prompt for MLD and 0.245s for \modelname.

\subsection{Re-usability of LEAD latent space}
We investigate the generalization capacity of our proposed LEAD latent projector on different motion latent diffusion models.

In particular, we consider a different diffusion model for text-to-motion generation such as MotionLCM~\cite{motionlcm}. MotionLCM offers speed enhancements by employing distillation, resulting in one-step (or few-step) inference. In Tab.~\ref{tab:motionlcm}, we report standard metrics to evaluate the performance of MotionLCM enhanced with the LEAD latent projector on the HumanML3D dataset. We observe that MotionLCM enhanced with the LEAD projector space improves in text-motion alignment (R-precision, MMdist) and motion quality (FID) consistently across different inference steps. We note here that we directly use the LEAD latent space \textbf{without retraining} for the MotionLCM model. Interestingly, while the LEAD latent space was trained using semantic language from CLIP, the learnt structure can be transferable to MotionLCM where the text is encoded using T5~\cite{t5} suggesting the robustness of the projector space.

\begin{table*}[h]
\small
\centering
\caption{\textcolor{red}{MotionLCM vs LEAD (MotionLCM) on the HumanML3D dataset. Bold indicates the best performance.}}
\label{tab:motionlcm}
\resizebox{\linewidth}{!}{%
\begin{tabular}{l||cccccccc}
\toprule
\multirow{2}{*}{Steps}  & \multirow{2}{*}{Model} & \multicolumn{1}{c}{FID ($\downarrow$)} & \multicolumn{3}{c}{R-precision ($\uparrow$)} & \multicolumn{1}{c}{MMdist ($\downarrow$)} & \multicolumn{1}{c}{Diversity ($\rightarrow$)} & \multicolumn{1}{c}{MModality ($\uparrow$)} \\
 \cmidrule(lr){3-5} \cmidrule(lr){6-6} \cmidrule(lr){7-7} \cmidrule(lr){8-8}
& &  & top1 & top2 & top3 &  &  &  \\
\midrule
\midrule
& Real & $0.002^{\pm .008}$ & $0.511^{\pm .003}$ & $0.703^{\pm .003}$ & $0.797^{\pm .002}$ & $2.974^{\pm .008}$ & $9.503^{\pm .065}$ & -  \\
\hline
% \bline
\multirow{1}{*}{1}& MotionLCM & $0.467^{\pm .012}$ & $0.502^{\pm .003}$ & $0.701^{\pm .002}$ & $0.803^{\pm .002}$ & $3.022^{\pm .009}$ & $\mathbf{9.631^{\pm .066}}$ & $\mathbf{2.172^{\pm .082}}$ \\
&  LEAD (MotionLCM)& $\mathbf{0.296^{\pm .007}}$ & \ $\mathbf{0.522^{\pm .003}}$ & $\mathbf{0.721^{\pm .003}}$ & $\mathbf{0.818^{\pm .003}}$& {$\mathbf{2.870^{\pm .010}}$ }& $9.820^{\pm .074}$ & {$1.974^{\pm .054}$} \\
\hline 
\multirow{1}{*}{2}& MotionLCM & $0.368^{\pm .011}$ & $0.505^{\pm .003}$ & $0.705^{\pm .002}$ & $0.805^{\pm .002}$ & $2.986^{\pm .008}$ & $\mathbf{9.640^{\pm .052}}$ & {$\mathbf{2.187^{\pm .094}}$} \\
& LEAD (MotionLCM)& $\mathbf{0.292^{\pm .008}}$ & $\mathbf{0.518^{\pm .003}}$ & $\mathbf{0.717^{\pm .003}}$ & $\mathbf{0.814^{\pm .002}}$ & $\mathbf{2.881^{\pm .009}}$ & $9.753^{\pm .090}$ & {$1.985^{\pm .049}$} \\
\hline 
\multirow{1}{*}{4}&  MotionLCM& $0.304^{\pm .012}$ & $0.502^{\pm .003}$ & $0.698^{\pm .002}$ & $0.798^{\pm .002}$ & $3.012^{\pm .007}$ & $\mathbf{9.607^{\pm .066}}$ & {$\mathbf{2.259^{\pm .092}}$} \\
&  LEAD (MotionLCM) & $\mathbf{0.268^{\pm .008}}$ & $\mathbf{0.509^{\pm .003}}$ & $\mathbf{0.708^{\pm .003}}$ & $\mathbf{0.806^{\pm .003}}$ & $\mathbf{2.932^{\pm .011}}$ & $9.682^{\pm .078}$ & {$2.110^{\pm .053}$} \\
\bottomrule
\end{tabular}}
\end{table*}

\subsection{Motion Textual Inversion}
\label{sec:mti_experiments}

To show the effectiveness of our proposed motion textual inversion we compare our approach (denoted $MTI_{proj}$) against the following baselines:

\begin{itemize}
\itemsep0em
    \item $MTI_{mld}$: Same as in image textual inversion~\cite{Gal:2022}, we first consider MTI on MLD, i.e. calculating the reconstruction in the VAE space, i.e.
    
    $||\epsilon -\epsilon_\theta(\vaeZ_t, t, \tau_\theta(y')||_2^2$. 
  
    \item $MTI_{feat}$:  Here we optimize using the reconstruction loss directly on the decoded motion, with $F$ defined in Eq~\ref{eq: F}: $\big\|\vaeD(\mathop{F}(\epsilon)) - \vaeD(\mathop{F}(\epsilon_\theta(z_t, t, \tau_\theta(y')))) \big\|_2^2 $
\end{itemize}

\noindent
For all variants, we evaluate FID, R-precision, MMdist and diversity on a mini-subset of 100 motions from HumanML3D. % (HumanML3D-mini). 
The results, as shown in Tab.~\ref{tab:mti}, demonstrate that MTI on the realigned space leads to improved results in terms of realism and multimodal alignment, compared to MTI on the MLD or motion feature space. Furthermore, we observe that incorporating the realignment mechanism during the text-to-motion generation with the newly learnt token proves beneficial in all MTI variants.

\begin{figure}[h]
    \centering
    \includegraphics[width=0.45 \textwidth]{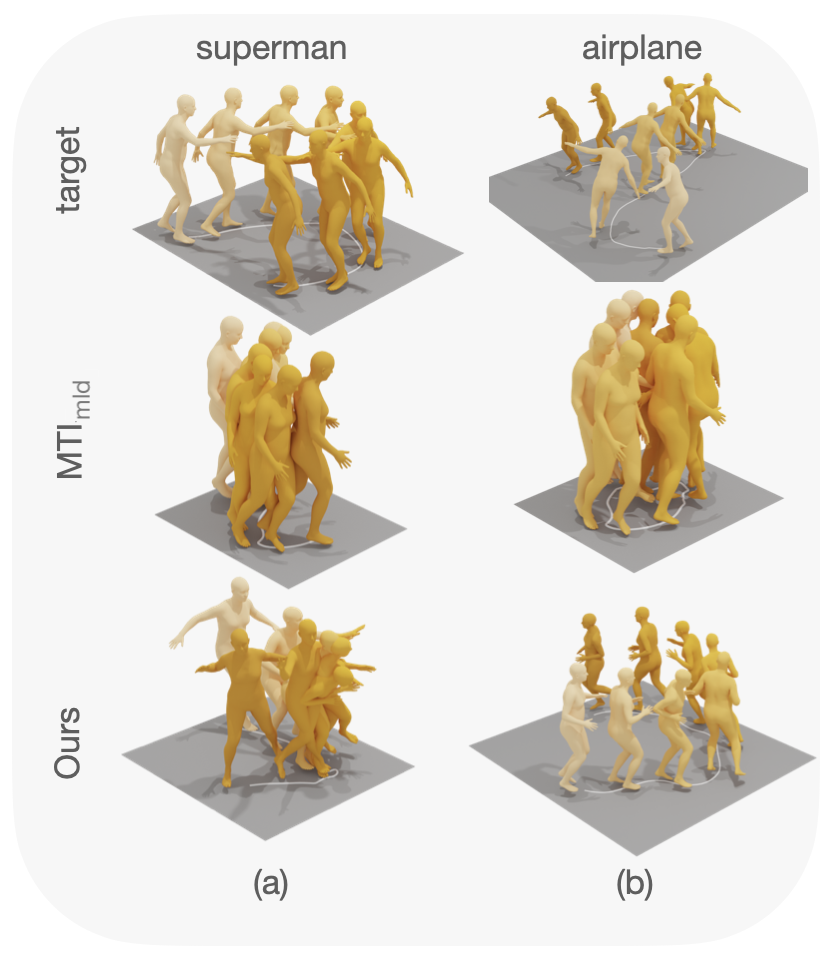}
    \caption{Results for MTI on out-of-distribution data, compared to the baseline without realignment.    
    Our results are more expressive and closer in content to the target examples.}
    \label{fig:mti_results}
\end{figure}

Additionally, we qualitatively evaluate on the 100styles dataset~\cite{Mason:2018} as it contains out-of-distribution sequences, and compare
ours ($\text{MTI}_{proj} \text{w/project}$) to the baseline $\text{MTI}_{MLD}$. \rev{We select distinctive styles from 100styles such as \textit{superman} and \textit{airplane} which were not seen during training, and learn a unique pseudo-word for each exemplar clip.} The results can be found in Fig.~\ref{fig:mti_results}, where we see that our generated motions are more expressive and closer in content to the target examples. On the other hand, $\text{MTI}_{MLD}$ (that is, optimizing directly towards the VAE latent) generates more inert motions, which illustrates the need for a realignment step during the inversion process. Animated results can be found in the supplementary video.

%%%%%%%%%
%%%%%%%%%

\begin{table}[t]
\caption{Ablation on projector module architecture and training losses. Top: transformer with long skip connections (skip) and a feedforward network (mlp). Bottom: we remove each of proposed losses to highlight their contribution.}
\label{tab:ablations1}

\resizebox{\linewidth}{!}{%
\begin{tabular}{l||HHccccc}
\toprule
\multirow{1}{*}{} & \multicolumn{3}{c}{R-prec. (top3) ($\uparrow$)} & \multicolumn{1}{c}{FID ($\downarrow$)} & \multicolumn{1}{c}{MMdist ($\downarrow$)} & \multicolumn{1}{c}{Div ($\rightarrow$)} & \multicolumn{1}{c}{MMod ($\uparrow$)} \\
% \cmidrule(lr){2-4} \cmidrule(lr){5-5} \cmidrule(lr){6-6} \cmidrule(lr){7-7} \cmidrule(lr){8-8}
% & top1 & top2 & top3 &  &  & &  \\
\midrule
\midrule
Real & $0.511\rmstd{^{\pm 0.003}}$ & $0.703\rmstd{^{\pm 0.003}}$ & $0.797\rmstd{^{\pm 0.002}}$ & $0.002\rmstd{^{\pm 0.008}}$ & $2.974\rmstd{^{\pm 0.008}}$ & $9.503\rmstd{^{\pm 0.065}}$ & - \\
\hline

\modelname (mlp) & 
$0.470\rmstd{^{\pm 0.003}}$ & 
$0.659\rmstd{^{\pm 0.002}}$ & 
$0.757\rmstd{^{\pm 0.002}}$ & 
$0.180\rmstd{^{\pm 0.010}}$ & 
$3.233\rmstd{^{\pm 0.010}}$ & 
$9.648\rmstd{^{\pm 0.077}}$ & 
$2.632\rmstd{^{\pm 0.108}}$ \\
\modelname (skip.) & 
$0.471\rmstd{^{\pm 0.003}}$ & 
$0.660\rmstd{^{\pm 0.003}}$ & 
$0.758\rmstd{^{\pm 0.002}}$ & 
$0.249\rmstd{^{\pm 0.007}}$ & 
$3.242\rmstd{^{\pm 0.007}}$ & 
$9.800\rmstd{^{\pm 0.091}}$ & 
$2.574\rmstd{^{\pm 0.084}}$ \\
\hline
\modelname - w/o KL &
0.479\rmstd{\textsuperscript{$\pm$0.003}} & \textbf{0.668\rmstd{\textsuperscript{$\pm$0.002}}} & \textbf{0.766\rmstd{\textsuperscript{$\pm$0.002}}} & 
0.320\rmstd{\textsuperscript{$\pm$0.010}} & 3.199\rmstd{\textsuperscript{$\pm$0.010}} & 9.799\rmstd{\textsuperscript{$\pm$0.086}} & 
2.489\rmstd{\textsuperscript{$\pm$0.094}} \\

\modelname - w/o Align &
0.472\rmstd{\textsuperscript{$\pm$0.003}} & 0.661\rmstd{\textsuperscript{$\pm$0.002}} & 0.759\rmstd{\textsuperscript{$\pm$0.002}} & 
0.156\rmstd{\textsuperscript{$\pm$0.007}} & 3.207\rmstd{\textsuperscript{$\pm$0.010}} & \textbf{9.509\rmstd{\textsuperscript{$\pm$0.079}}} & 
2.582\rmstd{\textsuperscript{$\pm$0.097}} \\

\modelname - w/o Rec &
0.438\rmstd{\textsuperscript{$\pm$0.003}} & 0.622\rmstd{\textsuperscript{$\pm$0.003}} & 0.727\rmstd{\textsuperscript{$\pm$0.002}} & 
0.630\rmstd{\textsuperscript{$\pm$0.014}} & 3.406\rmstd{\textsuperscript{$\pm$0.010}} & 8.888\rmstd{\textsuperscript{$\pm$0.069}} & 
\textbf{2.749\rmstd{\textsuperscript{$\pm$0.103}}} \\
\hline
\modelname  & 
$0.475\rmstd{^{\pm 0.003}}$ & 
$0.663\rmstd{^{\pm 0.004}}$ & 
$0.762\rmstd{^{\pm 0.003}}$ & 
$\mathbf{0.137}$ $\rmstd{^{\pm 0.006}}$ & 
$\mathbf{3.198\rmstd{^{\pm 0.011}}}$ & 
$9.656\rmstd{^{\pm 0.078}}$ & 
$2.572\rmstd{^{\pm 0.085}}$ \\
\bottomrule
\end{tabular}}
\end{table}

\subsection{Ablations}
\label{sec:ablation}

Finally, we conduct ablation studies on the architecture design for T2M on HumanML3D (Tab.~\ref{tab:ablations1}). In addition, we ablate the losses of \modelname for both T2M on HumanML3D (Tab.~\ref{tab:ablations1}), as well as MTI on HumanML3D-mini (Tab.~\ref{tab:ablations2}).

For T2M, we examine three different architectures (Tab.~\ref{tab:ablations1}): a simple MLP (first row), a transformer with long skip connections as in the VAE architecture of ~\cite{Chen:2023} (second row), and a transformer-based architecture (the proposed \modelname).
We observe that the best performance of \modelname consistently, across all metrics, is achieved when using a transformer-based architecture and all proposed losses (last row). 
Additionally,  
removing the reconstruction and KL loss leads to worse performance on FID and diversity while removing the alignment loss drops the R-precision.

For MTI, we examine the impact of the losses when using the projector without the KL loss (noKL), reconstruction loss (noREC) and alignment loss (noALIGN) and compare with the projector trained with all losses. Results in Tab.~\ref{tab:ablations2} show that the motion realism (FID) is significantly worse when the alignment loss is removed. We conclude that \modelname with all proposed losses achieves the best trade-off between R-precision, FID, MMdist and diversity.

\section{Conclusion}
\label{Sec:discussion}
We proposed \modelname, a human motion diffusion model equipped with a latent realignment mechanism.  
Leveraging the power of CLIP, we introduced a new latent space that is semantically better structured and can be trained with little effort given a pre-trained motion diffusion model.

Moreover, we proposed the new task of motion textual inversion, which optimizes for the textual embedding that best explains a set of example motions. This enables the generation of actions that are hard to explain with natural language, 
laying the groundwork for a more personalized content creation process.

    Our quantitative and qualitative results show improved performance on the text-to-motion task on the HumanML3D and KIT-ML datasets, achieving greater realism without compromising diversity. Our evaluation of motion textual inversion on the 100styles dataset further shows that the realigned latent space is \rev{more suitable for the generation of out-of-distribution sequences learned with few examples}. We believe this is a promising direction, as it opens new opportunities for intuitive, personalized motion generation with only a few exemplar motions.

\noindent
\textbf{Limitations}
As demonstrated quantitatively and reflected by the user study, compared to the baselines, motions generated with \modelname are more realistic. However, they still sometimes exhibit motion artifacts such as foot-sliding. Furthermore, for long and sequential textual descriptions \modelname may struggle to produce motions that faithfully comply to the motion description. Finally, even though \modelname can generate arbitrary-length motions due to its design, it remains bounded by the maximum motion length present in the dataset.

\noindent
\textbf{Social Impact} Personalized motion generation from text introduces the risk of misuse in developing deceptive artificial content and raises ethical concerns regarding privacy and consent, as the technology may create motions that portray a specific person without explicit authorization from them.

\bibliographystyle{eg-alpha-doi}
\bibliography{main, shortlists}

%-------------------------------------------------------------------------
\newpage

\end{document}